%% file: main.tex
\documentclass{article}
\usepackage{spconf,amsmath,graphicx}
\usepackage{hyperref}
\usepackage{url}
\usepackage{tabu}
\usepackage{enumitem}
\usepackage{subcaption}
\usepackage{adjustbox}
\usepackage{booktabs} % Table lines

\title{YouTube SFV+HDR Quality Dataset}

\name{Yilin Wang, Joong Gon Yim, Neil Birkbeck, Balu Adsumilli}
\address{Google Inc., 1600 Amphitheatre Pkwy., Mountain View, CA, USA 94043}

\begin{document}

\maketitle

\begin{abstract}
The popularity of Short form videos (SFV) has grown dramatically in the past few years, and has become a phenomenal video category with billions of viewers. Meanwhile, High Dynamic Range (HDR) as an advanced feature also becomes more and more popular on video sharing platforms. As a hot topic with huge impact, SFV and HDR bring new questions to video quality research: 1) is SFV+HDR quality assessment significantly different from traditional User Generated Content (UGC) quality assessment? 2) do objective quality metrics designed for traditional UGC still work well for SFV+HDR? To answer the above questions, we created the first large scale SFV+HDR dataset with reliable subjective quality scores, covering 10 popular content categories. Further, we also introduce a general sampling framework to maximize the representativeness of the dataset. We provided a comprehensive analysis of subjective quality scores for Short form SDR and HDR videos, and discuss the reliability of state-of-the-art UGC quality metrics and potential improvements.
\end{abstract}

\input{intro}

\input{dataset_creation}

\input{subjective_data_analysis}

\input{objective_metric_performance}

\input{conclusion}

\bibliographystyle{IEEEbib}
\bibliography{citations}

\end{document}

%% file: intro.tex
\section{Introduction}
\label{sec:intro}
Compared to long form videos, Short form videos (SFV) are quicker to view and consume, better aligned with our fast-paced lives, which made it a phenomenal video category in past few years. Understanding SFV quality is an important topic, which benefits many areas including video creation, compression, transmission, search, and recommendation. Meanwhile, High Dynamic Range (HDR) becomes a more and more popular video feature, which has been widely supported by recent devices. How valuable HDR is on User Generated Content (UGC) especially SFV is also an interesting topic. Besides real HDR, Standard Dynamic Range (SDR) converted from HDR (noted as HDR2SDR) is also an important video category with even more consumers. Is there a significant quality difference between native SDR and HDR2SDR? If yes, more video creators would be encouraged to generate HDR contents. To answer these questions, we need a representative SFV+HDR dataset.

However, there are limited public resources for SFV and HDR contents, especially large scale datasets. 
SVD~\cite{Jiang2019SVDdataset} collected 500K SFV URLs from Douyin, with labeled pairs of near-duplicate videos. AutoShot~\cite{Zhu2023Autoshot} provided about 1000 SFV with shot boundary annotations. MMSVD-Douyin~\cite{Zhang2023MMSVD-Douyin} included 4684 SFV with numbers of ``likes'', ``shares'', and ``comments'', but the data are not publicly available yet.
NTIRE 2021 HDR Challenge~\cite{Perez2021NTIRE2021HDR} prepared 1500 HDR samples for model training, but the data is only available for registered participants. None of these datasets provided subjective quality labels. Regarding video quality research, existing large scale UGC datasets (e.g. KoNViD-1k~\cite{konvid1k}, LIVE-VQC~\cite{Sinno12018CrowdsourcedQualityDataset}, YouTube UGC~\cite{Wang2019UGCDataset}, Youku-V1K~\cite{Xu2021YoukuV1K}, and LSVQ~\cite{Ying2021PatchVQ}) mainly focused on landscape and long form videos. YouTube UGC dataset~\cite{Wang2019UGCDataset} had about 100 vertical videos, but still sampled from long form videos. 
% Our dataset is the first large scale dataset focusing on Short quality.

As the largest video sharing platform in the world, YouTube is an ideal source to sample SFV contents. To facilitate more research on SFV and HDR, we created the first large scale YouTube SFV+HDR quality dataset. The key characteristics of this dataset are summarized in Table~\ref{tab:CharacteristicsoftheYouTubeShortsDataset}. 
Videos and subjective data are available
on\\ \href{https://media.withyoutube.com/sfv-hdr}{\textit{https://media.withyoutube.com/sfv-hdr}}.

\begin{table}[ht]
\small
    \centering
    \begin{tabular}{| l || c |}
        \hline
        \textbf{Color Space} & SDR, HDR   \\\hline
        \textbf{Resolution}          &    $1080\times{1920}$       \\\hline
        \textbf{Video length} & 5s  \\\hline
        \textbf{Content category}      & Animal, Cooking, Dance, Gameplay, Health,  \\
                             &   Hobby, Music, Society, Speech, Sports \\\hline
        % \textbf{Initial pool size} & SDR (80000, in late 2023), HDR (4000) \\  \hline
        \textbf{Videos} & SDR (2030), HDR2SDR (2000), HDR (2000) \\  \hline
        \textbf{Subjective scores} & SDR (2030), HDR2SDR (2000), HDR (300) \\  \hline
    \end{tabular}
    \caption{Characteristics of the YouTube SFV+HDR dataset.}
    \label{tab:CharacteristicsoftheYouTubeShortsDataset}
\end{table}
% \\
The contributions of this paper are as follows:
% \begin{itemize}[leftmargin=.15in] 
% \setlength\itemsep{0.0em}
% \vspace{-0.2cm}
\begin{enumerate}[nosep]
\item A public dataset with 4030 SFV contents (2030 SDR and 2000 HDR) and corresponding subjective scores, covering 10 popular SFV content categories.
\item A three-step sampling framework to maximize the representativeness of videos (Section~\ref{sec:Dataset Creation}).
\item A comprehensive analysis of subjective quality of SFV in SDR, HDR2SDR, and HDR (Section~\ref{sec:Subjective Data Analysis}).
\item Evaluation of SOTA UGC quality models on SFV quality, and potential directions for improvement, e.g. Gameplay and HDR2SDR SFV. (Section~\ref{sec:Objective Metric Performance}).
\end{enumerate}
% \end{itemize}

%% file: dataset_creation.tex
\section{Three Step video Sampling Framework}
\label{sec:Dataset Creation}

%Creating a large scale quality dataset is not an easy task, and the ``quality'' of the dataset could be affected by various practical constraints. 
Care must be taking when sampling a large-scale video dataset, 
as the the ``quality'' of the dataset could be affected by various practical constraints. 
Naively including all available samples in the dataset may not be a good idea, and is usually impractical. A core step of creating a quality dataset is the subjective test, which is expensive and time consuming. In most cases, it is unaffordable to run subjective tests on all available samples.
Given the limited budget for subjective tests, another straightforward approach is random sampling. However, the raw set usually contains many duplicates or very similar data, which should not be sampled more than once. Random sampling cannot reduce such data redundancy, and may fail to represent relatively minor categories. A more sophisticated sampling strategy is highly desired, which usually requires domain specific knowledge.
Three core practical considerations that need to be addressed in methodology used to create a video quality dataset are the following: 
\begin{enumerate}[nosep]
    \item The identification of a representative sampling pool.
    \item A fair sampling method that meaningfully covers the entire space.
    \item Maximizing the data representativeness/diversity using a fixed size.
\end{enumerate} 
% Here we listed three core questions that need to be addressed when creating a quality dataset:
% \begin{enumerate}[nosep]
%     \item How to identify a representative sampling pool?
%     \item How to fairly sample the entire space?
%     \item How to maximize the data representativeness/diversity given the fixed size of the dataset?
% \end{enumerate} 
% \\
To answer these questions, we proposed a general framework to separate the creation procedure into three steps: sampling pool construction, feature space sampling, and final content review. 
In following sections, we will use the YouTube SFV+HDR quality dataset as an example to discuss each step in details.

\subsection{Sampling Pool Construction}
\label{subsec:Sampling Pool Construction}
Our raw video pool ideally includes all YouTube SFV with the Creative Commons license. However, it doesn't mean we can cover all video characteristics with sufficient samples (e.g. 10 sampling points per dimension) in one dataset. Common video characteristics include content, original quality, video (spatial/temporal) complexity, color space, resolution, frame rate, video length, freshness, etc. It is preferable to identify key  characteristics of the study, and remove less important dimensions. Regarding a SFV quality dataset, we think resolution, frame rate, and video length are less important for exploring distinct characteristics of SFV. Thus we chose the most popular settings for resolution and frame rate, which are $1080\times{1920}$ and 30FPS respectively. For video length, we cropped the first 5 seconds of the entire video, which fairly represented the quality for most SFV.
To sufficiently represent the current trend of SFV, we selected 80,000 recently uploaded SDR videos with Creative Commons license.
Video content is another important dimension to study human opinions on SFV quality. We selected 10 popular SFV categories annotated by Knowledge Graph~\cite{Singhal12KnowledgeGraph}: Animal, Cooking, Dance, Gameplay, Health, Hobby, Music, Society, Speech, and Sports. In this way, our initial SDR pool has 10 subsets, corresponding to 10 content categories, and each subset contains 1000 to 7000 samples.

HDR is another focus of our dataset, and has additional constraint.
%which turned out to be in a different situation. 
Compared to SDR SFV, the total number of HDR SFV with Creative Common license is smaller, so we included all available 1080P HDR SFV in the initial HDR pool, still cropped the first 5s. Only part of HDR SFV were associated with above mentioned content labels, and others' content categories are labeled as ``unknown''. 

The initial SDR and HDR sampling pools were created as outlined above.  Although they are in different situations in terms of the pool size, we still followed the same creation rationale, i.e. maintaining key characteristics of the interest and removing less important dimensions to reduce noise.

\subsection{Feature Space Sampling}
\label{subsec:Feature Space Sampling}
The identified sampling pool usually contains orders of magnitudes more videos than the target size. In our case, the target size for SDR SFV dataset is 2000, while the size of the SDR pool is 80000. How to fairly sample videos is another important topic. Purely random sampling has a risk to be biased with the current data distribution and poorly represent minor categories. To better represent the entire set, we divided the sampling space by three basic video features: spatial information (SI), temporal information (TI), and perceptual quality. Here we followed ITU-T Rec. P910~\cite{ITU-T-RecP910} to calculate SI and TI. For perceptual quality, the precise value requires subjective tests, which is not available at this stage. Alternatively we used UVQ (old name CoINVQ)~\cite{Wang2021Coinvq} to approximate perceptual quality, which should be sufficient for the sampling purpose.
% We further investigated the distributions for individual content categories, and observed that some categories have very different distortions than others.

\begin{figure}
\centering
\includegraphics[width=0.45\textwidth, trim={0cm 0cm 0cm 0cm}, clip]{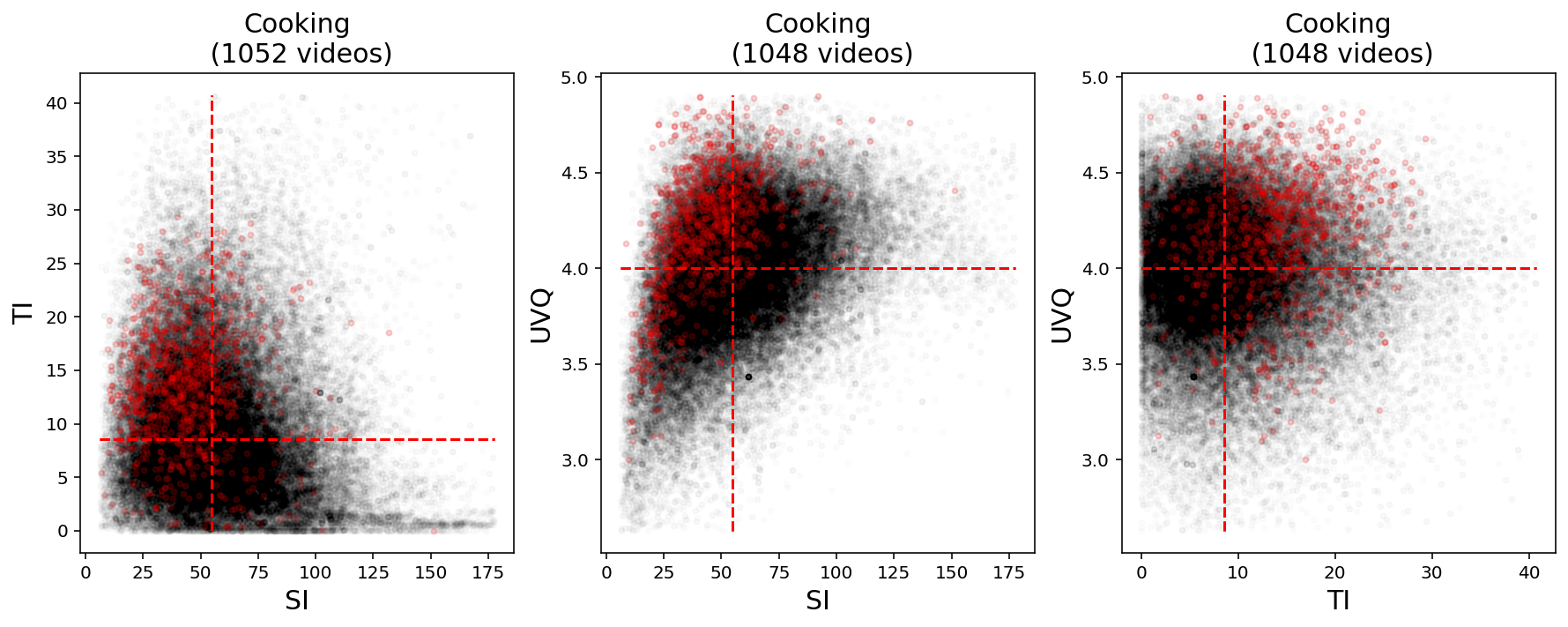}\\
\includegraphics[width=0.45\textwidth, trim={0cm 0cm 0cm 0cm}, clip]{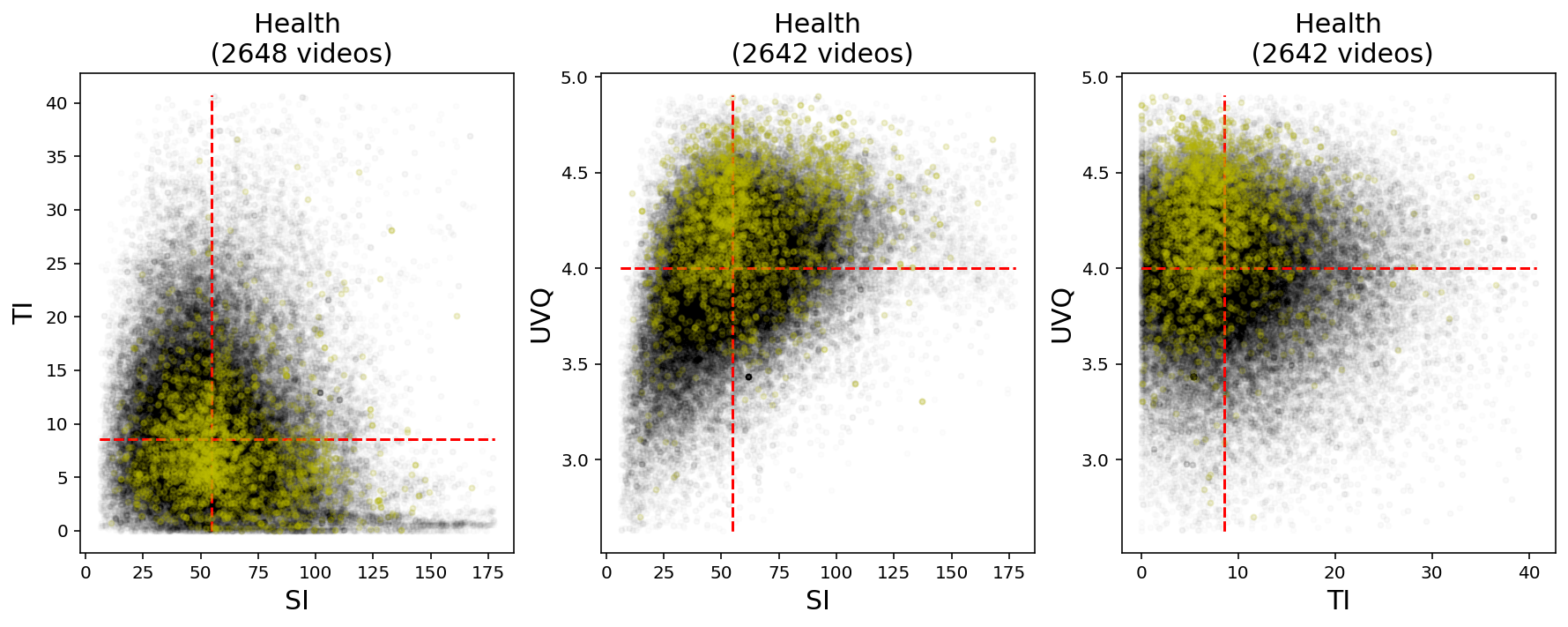}\\
\includegraphics[width=0.45\textwidth, trim={0cm 0cm 0cm 0cm}, clip]{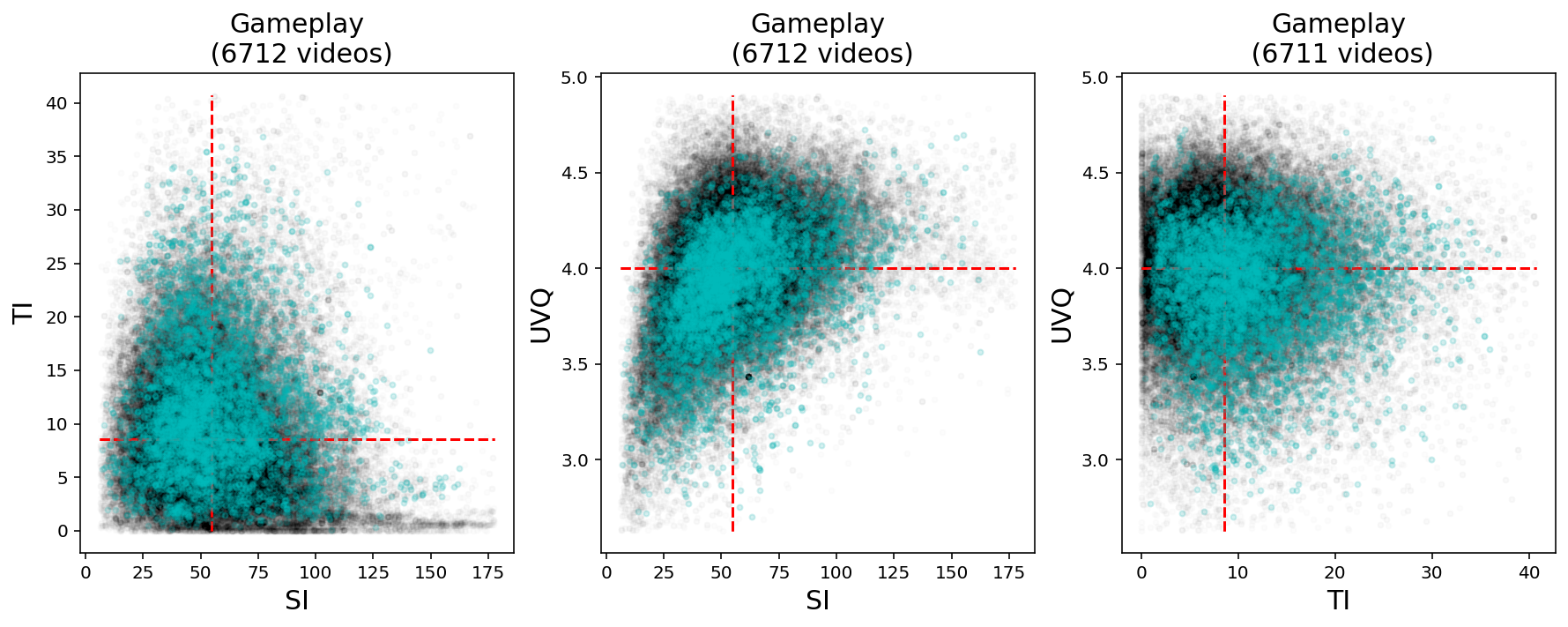}\\
\caption{Distributions of SI, TI, and UVQ for the entire pool (black) and three content categories Cooking (red), Health (yellow), and Gameplay (blue), whose distributions are significantly different from one another}
\label{fig:distribution_sitiuvq}
\end{figure}

\begin{figure*}
  \begin{subfigure}{0.18\linewidth}
    \centering
    \includegraphics[width=0.48\linewidth, trim={0cm 0cm 0cm 0cm}, clip]{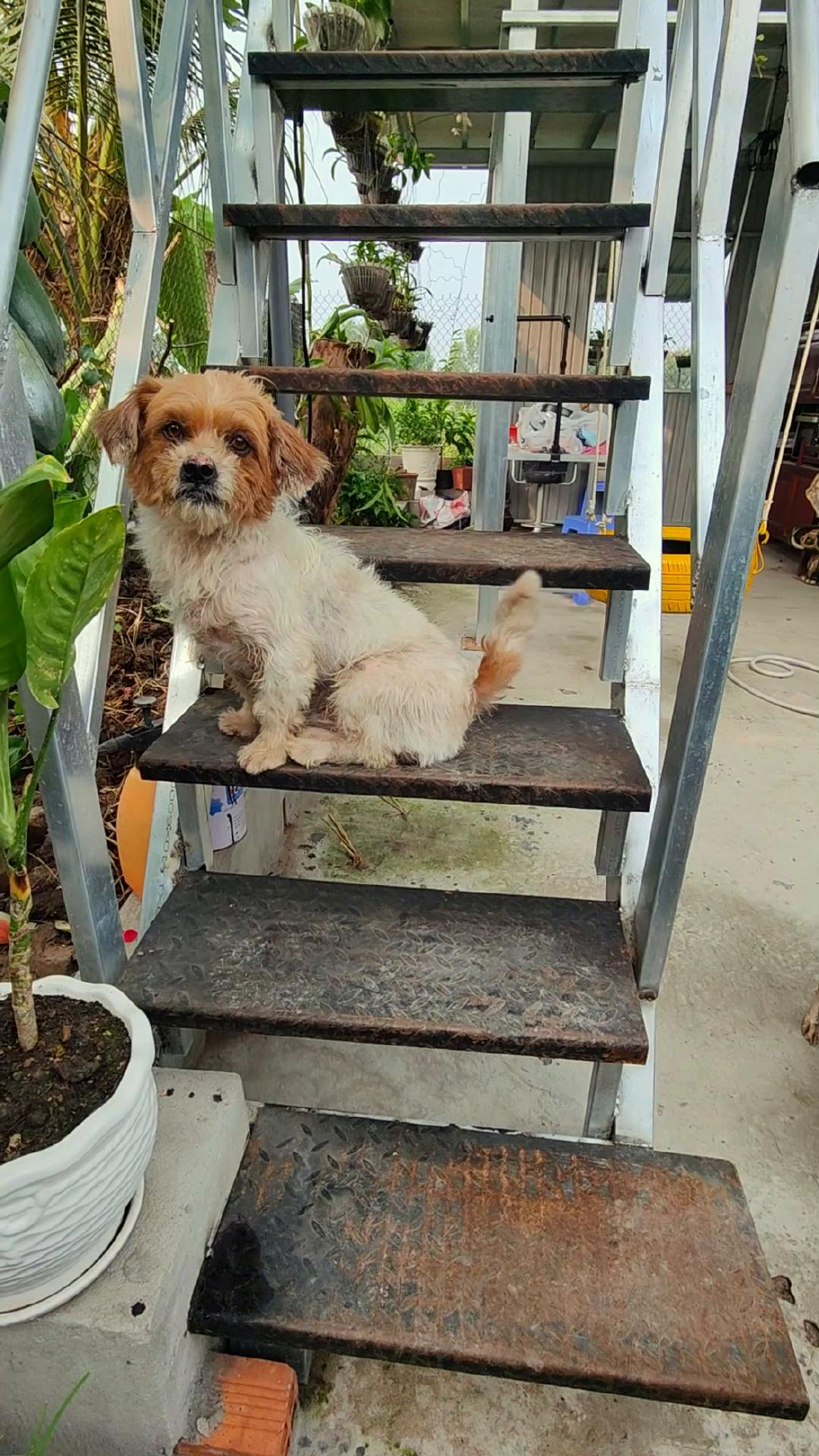}
    \includegraphics[width=0.48\linewidth, trim={0cm 0cm 0cm 0cm}, clip]{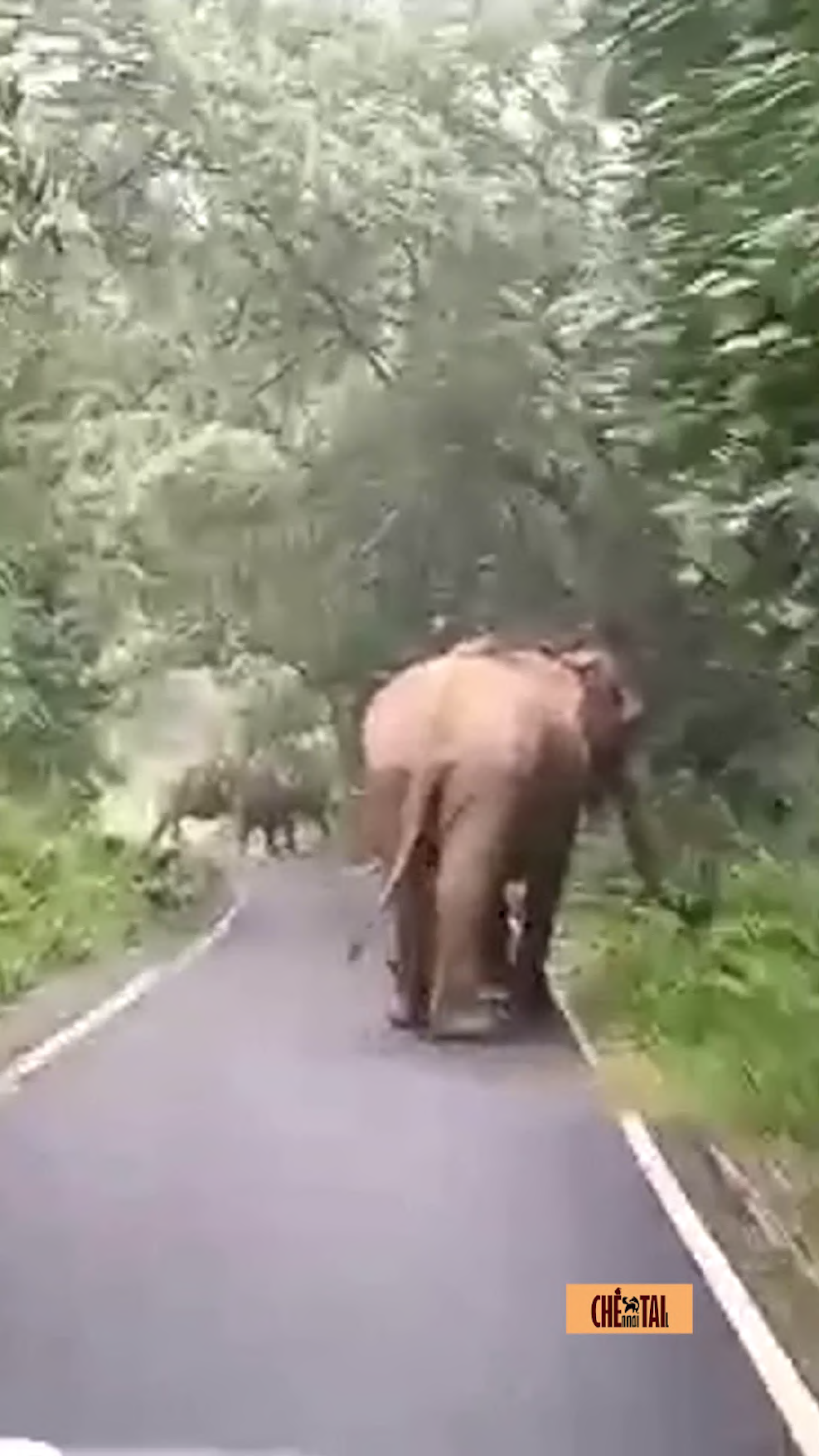}
    \captionsetup{labelformat=empty}
    \caption{Animal}
  \end{subfigure}
  \hspace{0.01 in}
  \begin{subfigure}{0.18\linewidth}
    \centering
    \includegraphics[width=0.48\linewidth, trim={0cm 0cm 0cm 0cm}, clip]{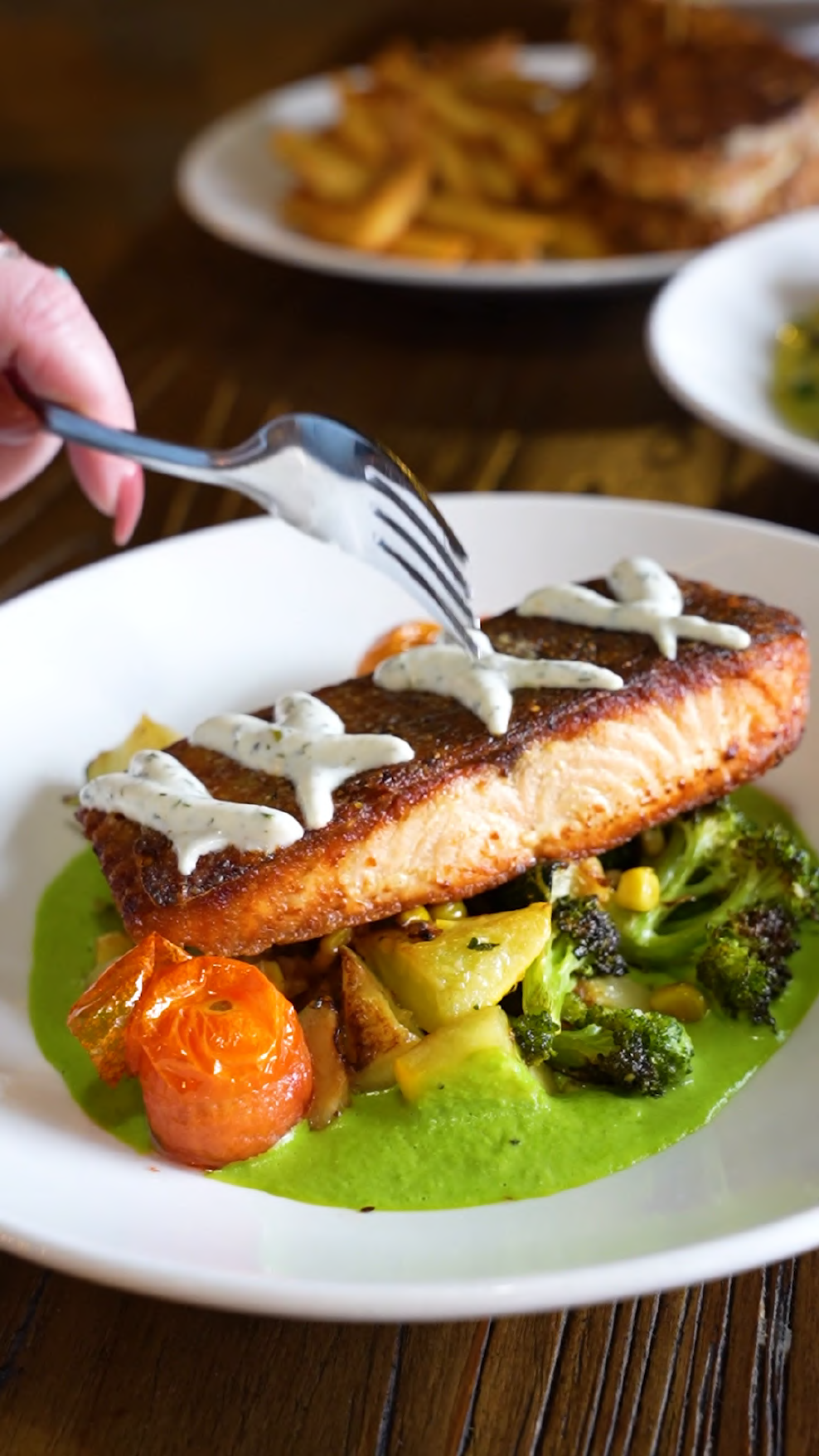}
    \includegraphics[width=0.48\linewidth, trim={0cm 0cm 0cm 0cm}, clip]{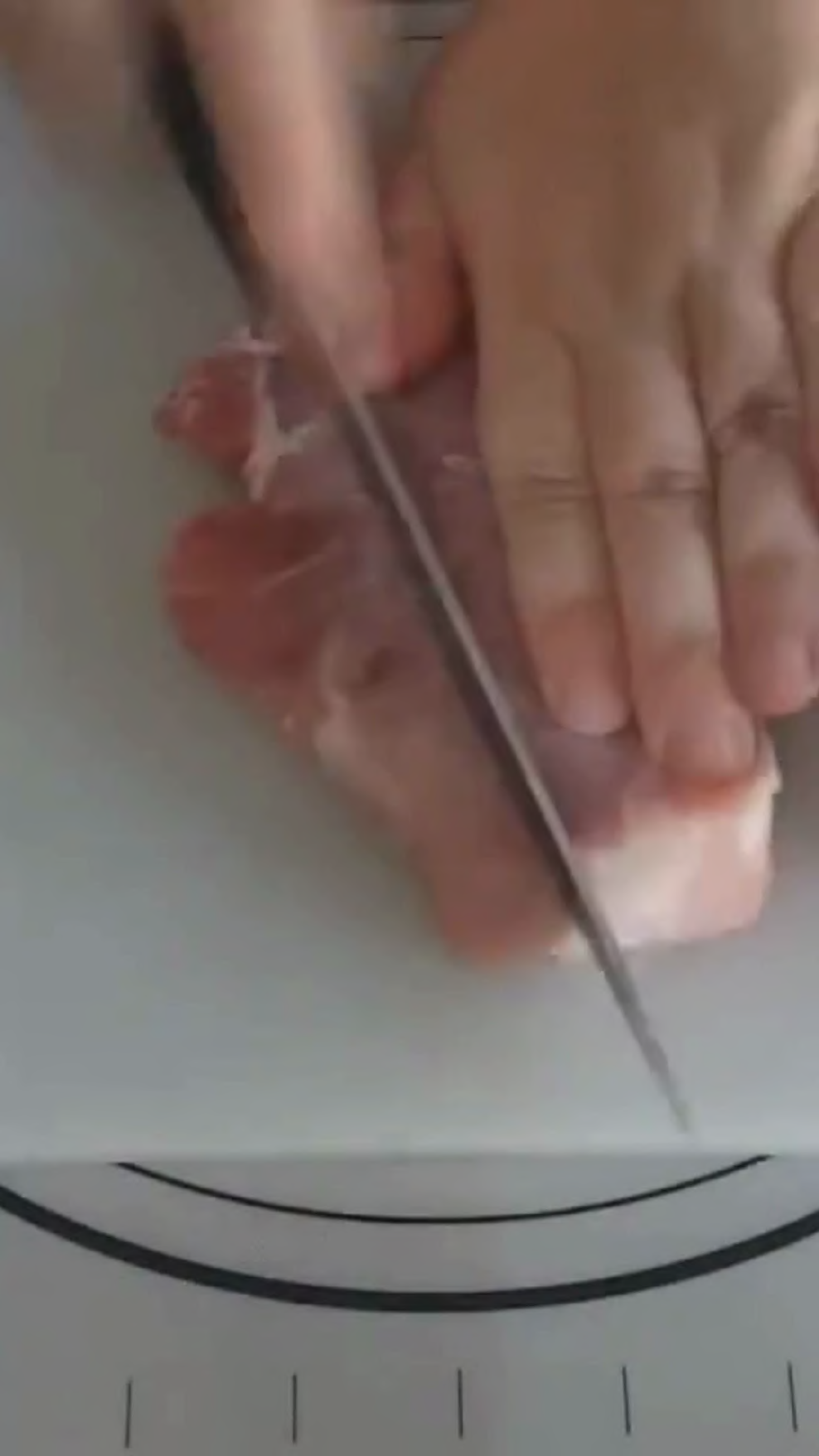}
    \captionsetup{labelformat=empty}
    \caption{Cooking}
  \end{subfigure}
  \hspace{0.01 in}
  \begin{subfigure}{0.18\linewidth}
    \centering
    \includegraphics[width=0.48\linewidth, trim={0cm 0cm 0cm 0cm}, clip]{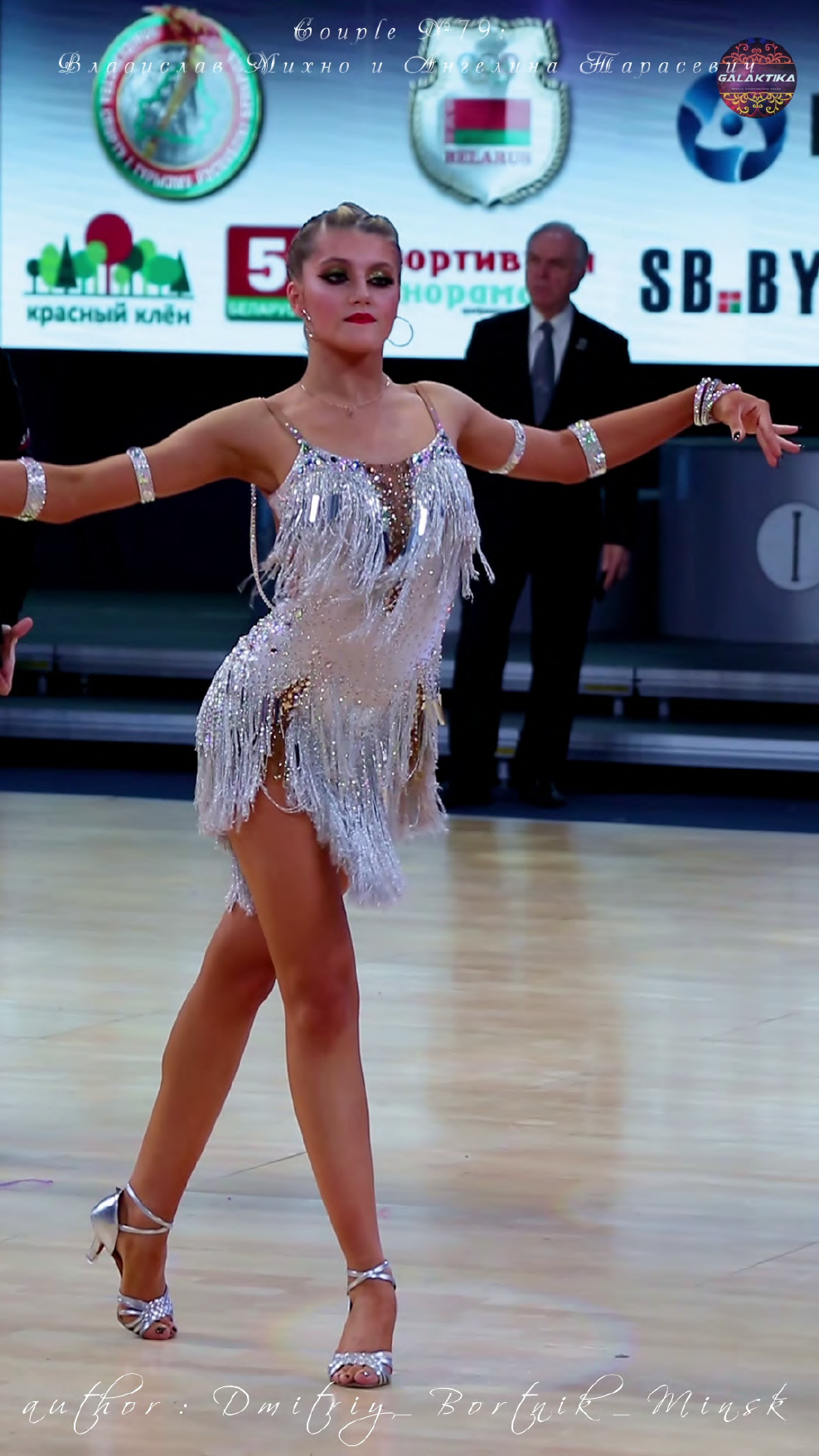}
    \includegraphics[width=0.48\linewidth, trim={0cm 0cm 0cm 0cm}, clip]{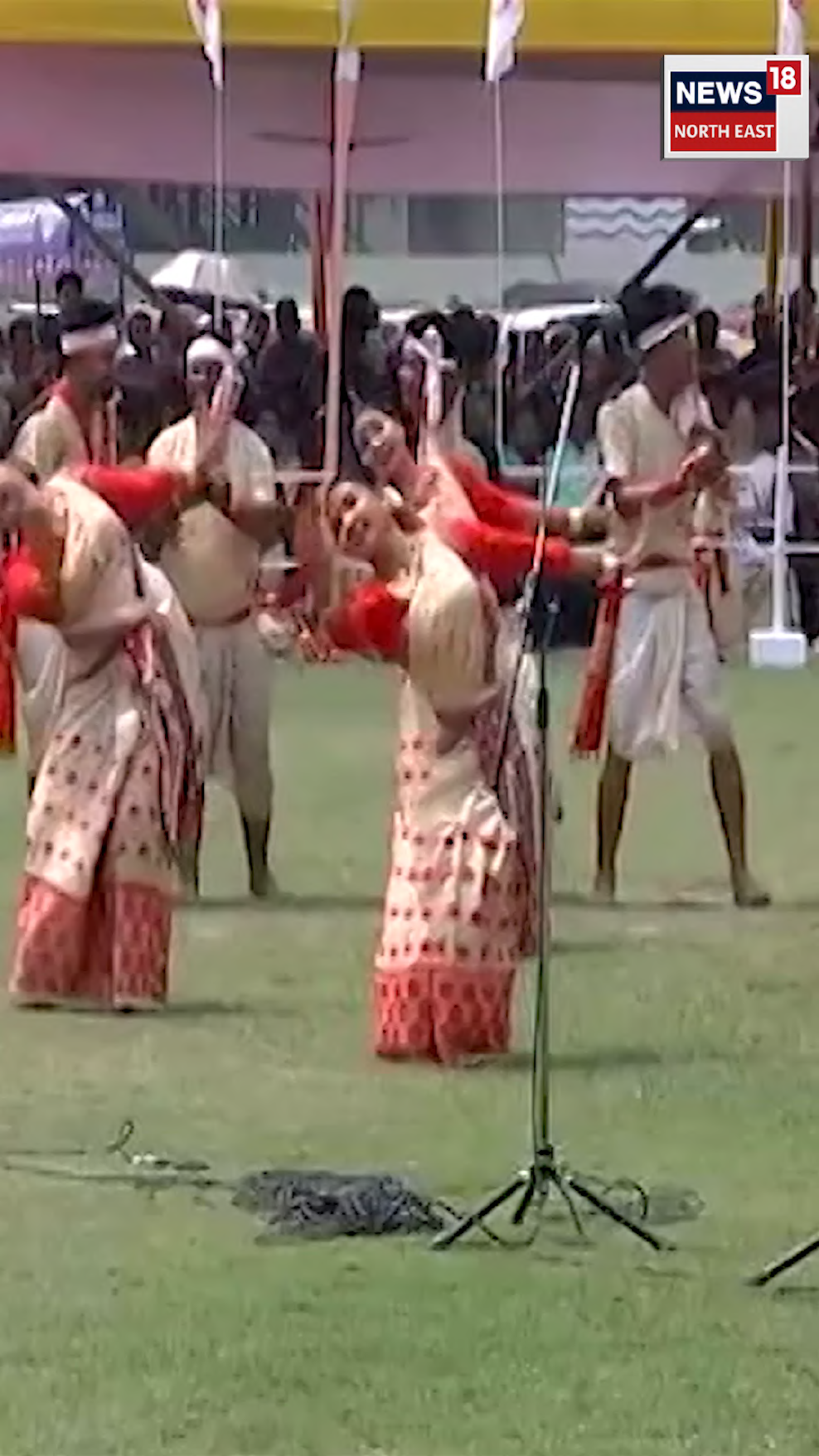}
    \captionsetup{labelformat=empty}
    \caption{Dance}
  \end{subfigure}
  \hspace{0.01 in}
  \begin{subfigure}{0.18\linewidth}
    \centering
    \includegraphics[width=0.48\linewidth, trim={0cm 0cm 0cm 0cm}, clip]{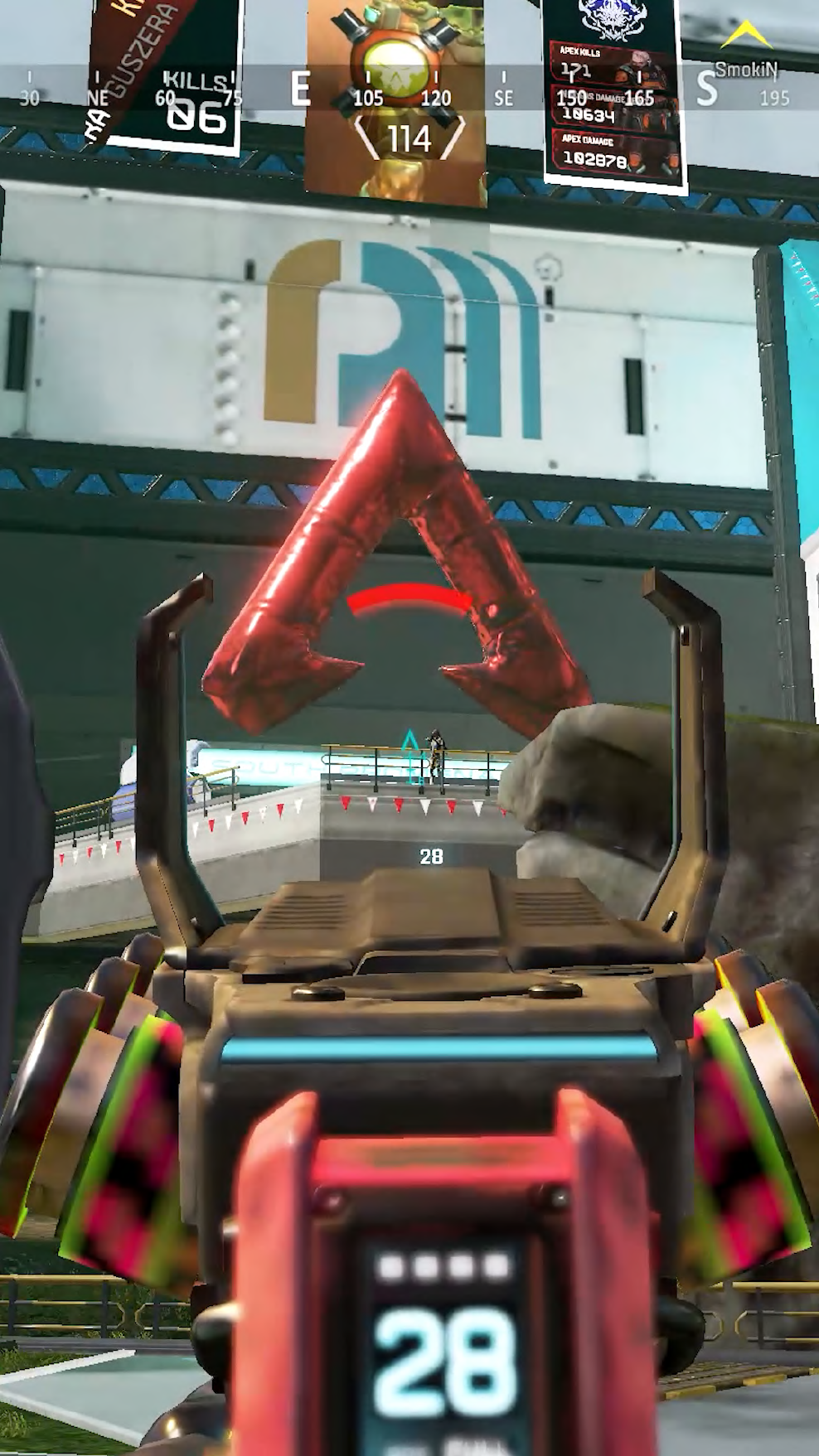}
    \includegraphics[width=0.48\linewidth, trim={0cm 0cm 0cm 0cm}, clip]{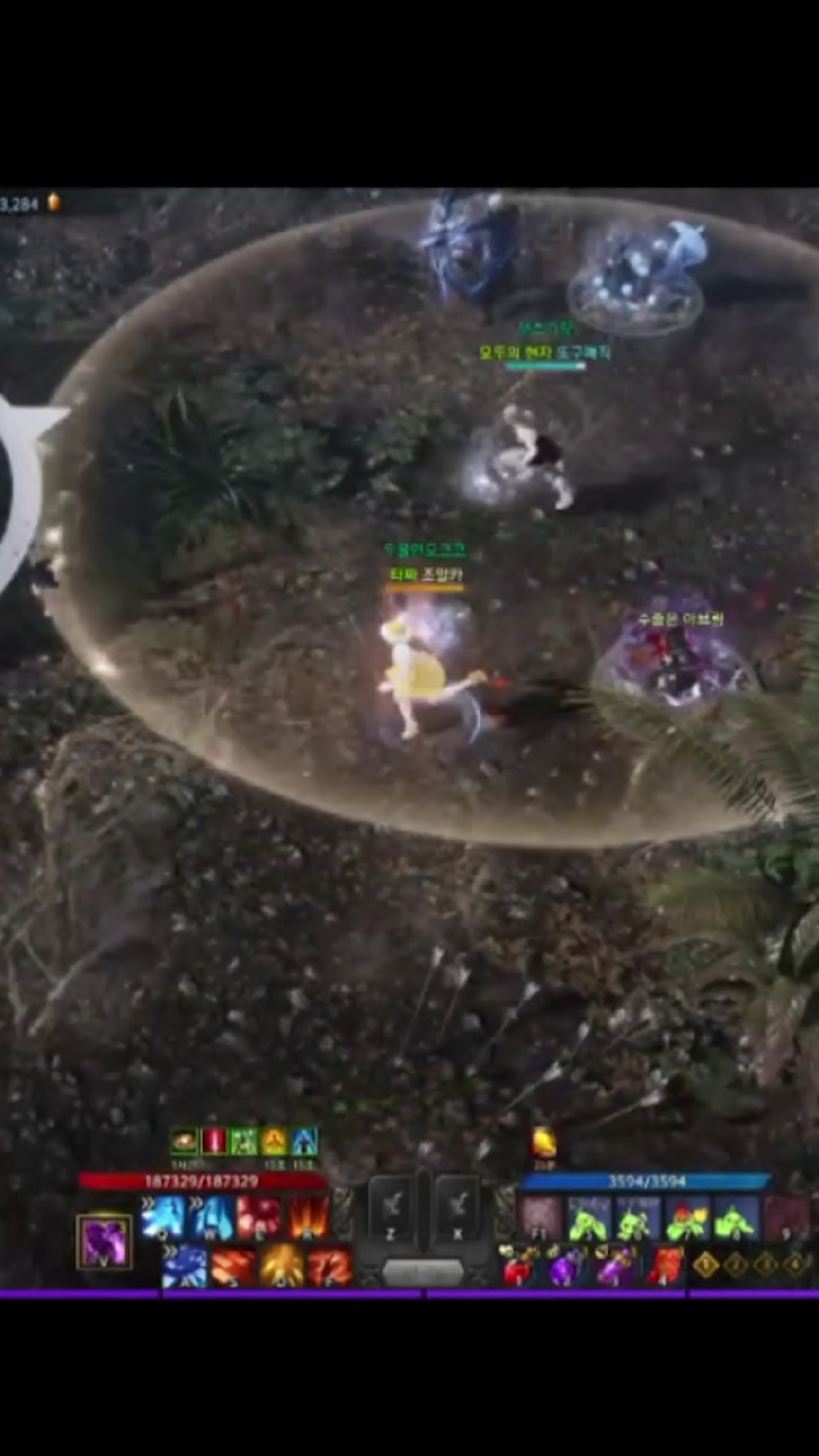}
    \captionsetup{labelformat=empty}
    \caption{Gameplay}
  \end{subfigure}
  \hspace{0.01 in}
  \begin{subfigure}{0.18\linewidth}
    \centering
    \includegraphics[width=0.48\linewidth, trim={0cm 0cm 0cm 0cm}, clip]{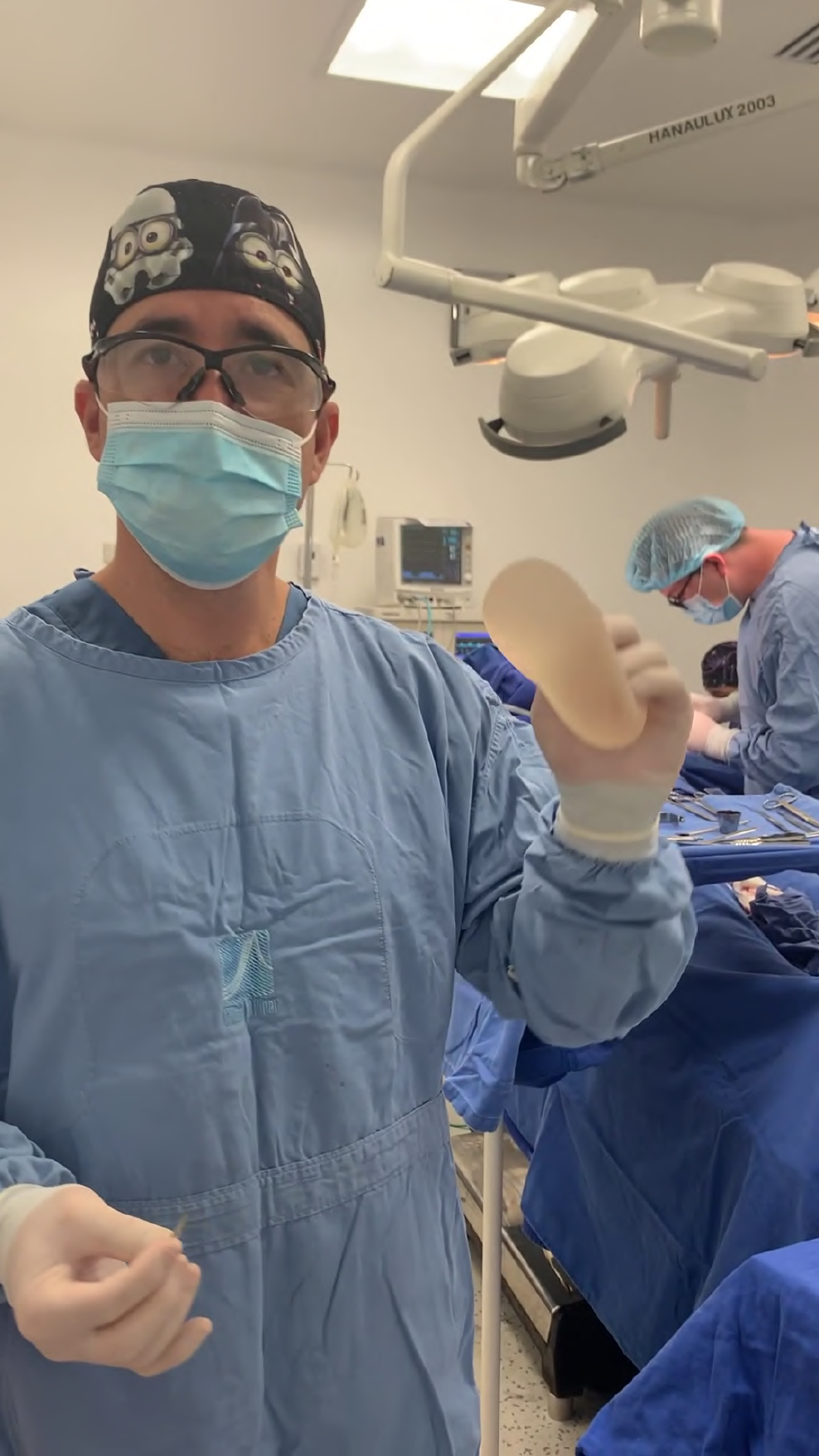}
    \includegraphics[width=0.48\linewidth, trim={0cm 0cm 0cm 0cm}, clip]{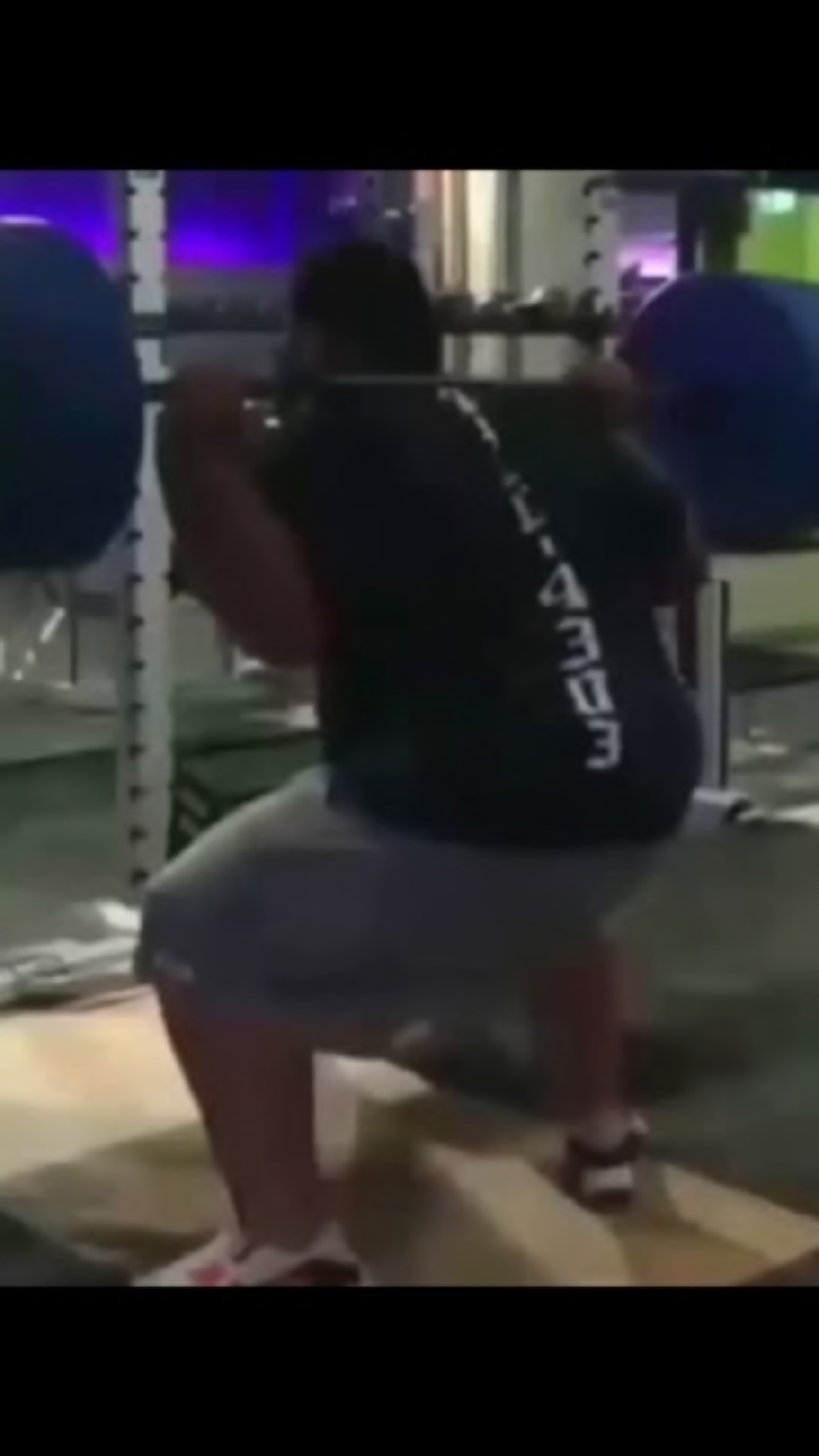}
    \captionsetup{labelformat=empty}
    \caption{Health}
  \end{subfigure}
\\
\\
\\
  \begin{subfigure}{0.18\linewidth}
    \centering
    \includegraphics[width=0.48\linewidth, trim={0cm 0cm 0cm 0cm}, clip]{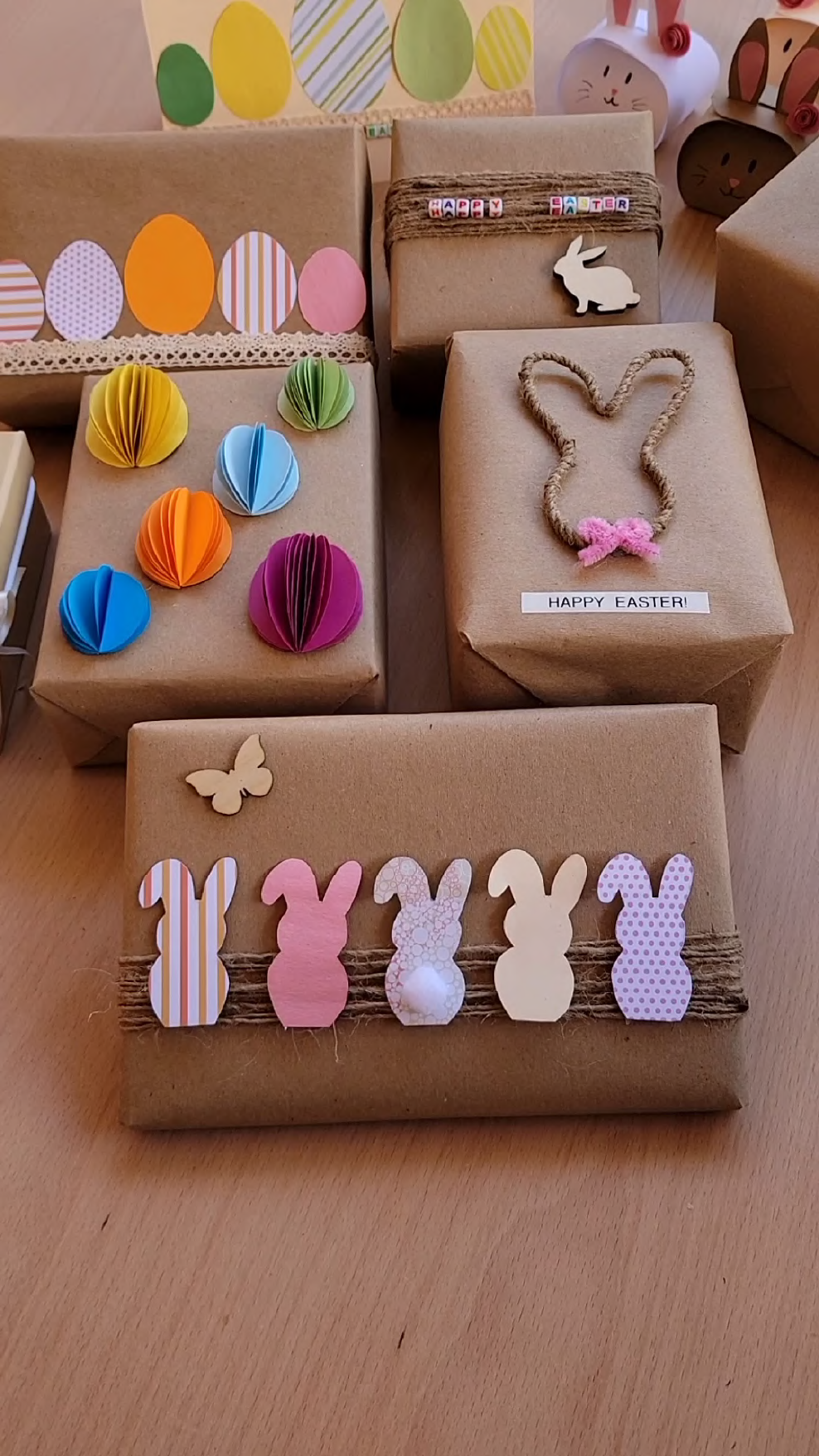}
    \includegraphics[width=0.48\linewidth, trim={0cm 0cm 0cm 0cm}, clip]{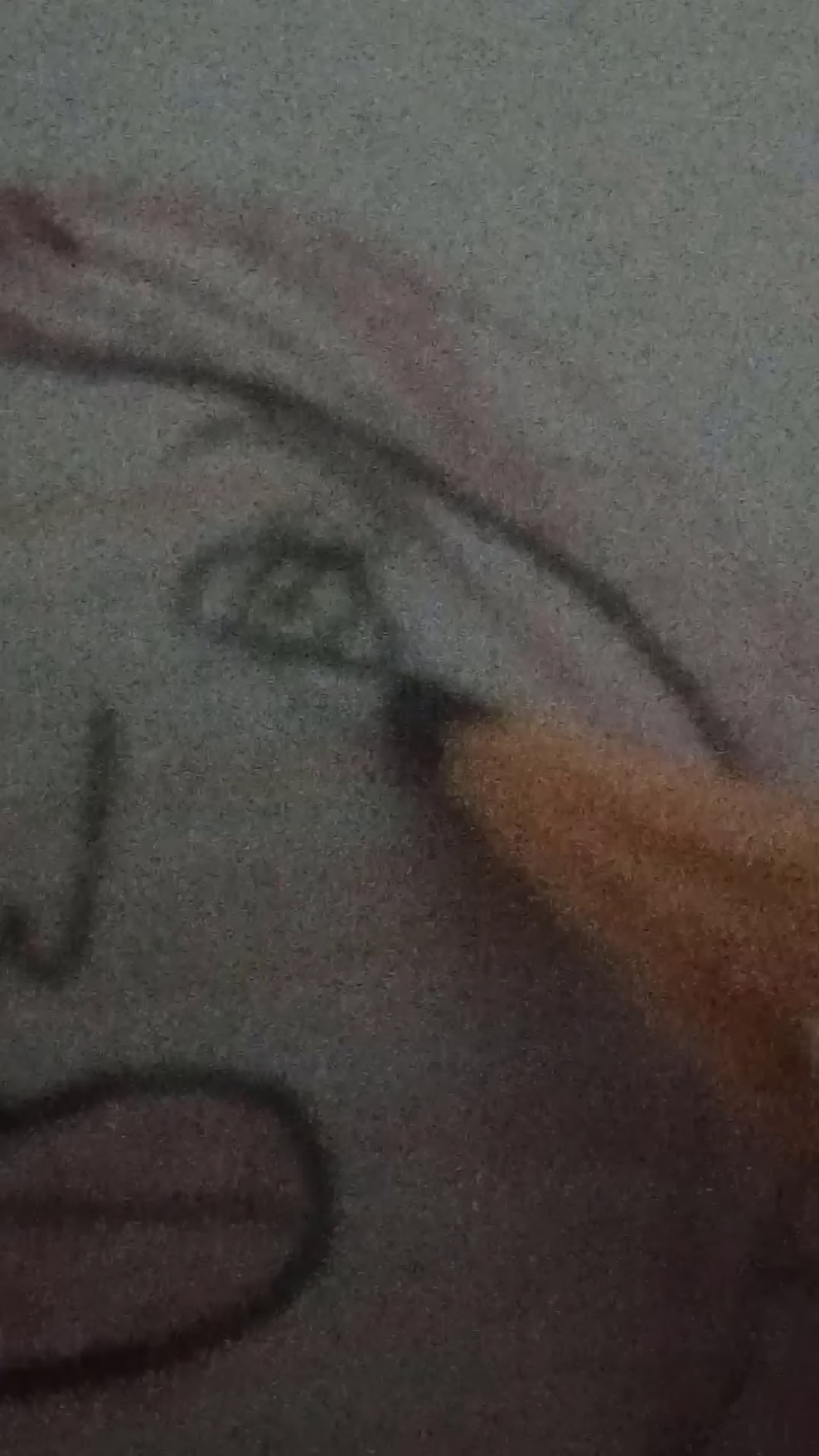}
    \captionsetup{labelformat=empty}
    \caption{Hobby}
  \end{subfigure}
  \hspace{0.01 in}
  \begin{subfigure}{0.18\linewidth}
    \centering
    \includegraphics[width=0.48\linewidth, trim={0cm 0cm 0cm 0cm}, clip]{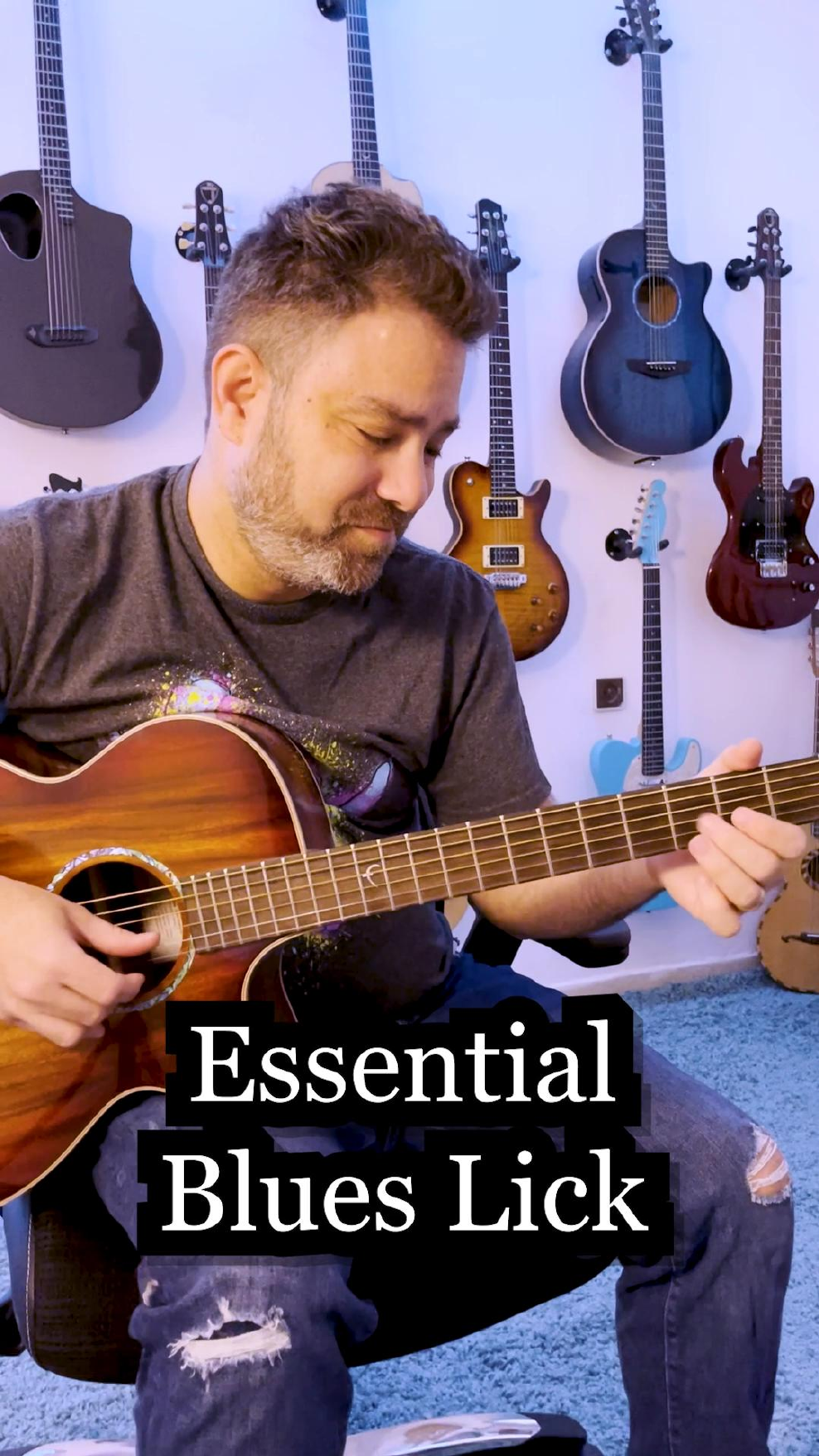}
    \includegraphics[width=0.48\linewidth, trim={0cm 0cm 0cm 0cm}, clip]{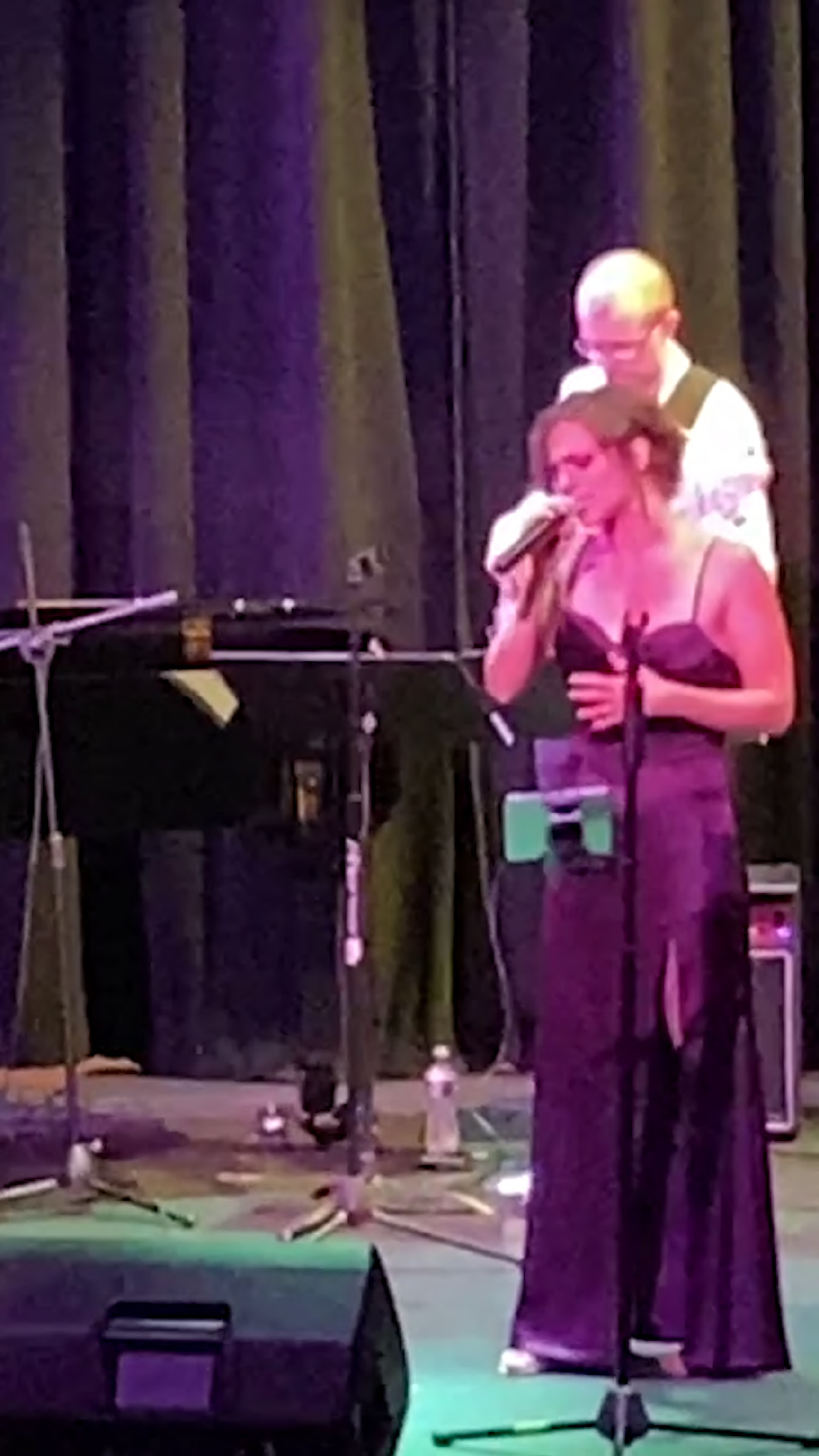}
    \captionsetup{labelformat=empty}
    \caption{Music}
  \end{subfigure}
  \hspace{0.01 in}
  \begin{subfigure}{0.18\linewidth}
    \centering
    \includegraphics[width=0.48\linewidth, trim={0cm 0cm 0cm 0cm}, clip]{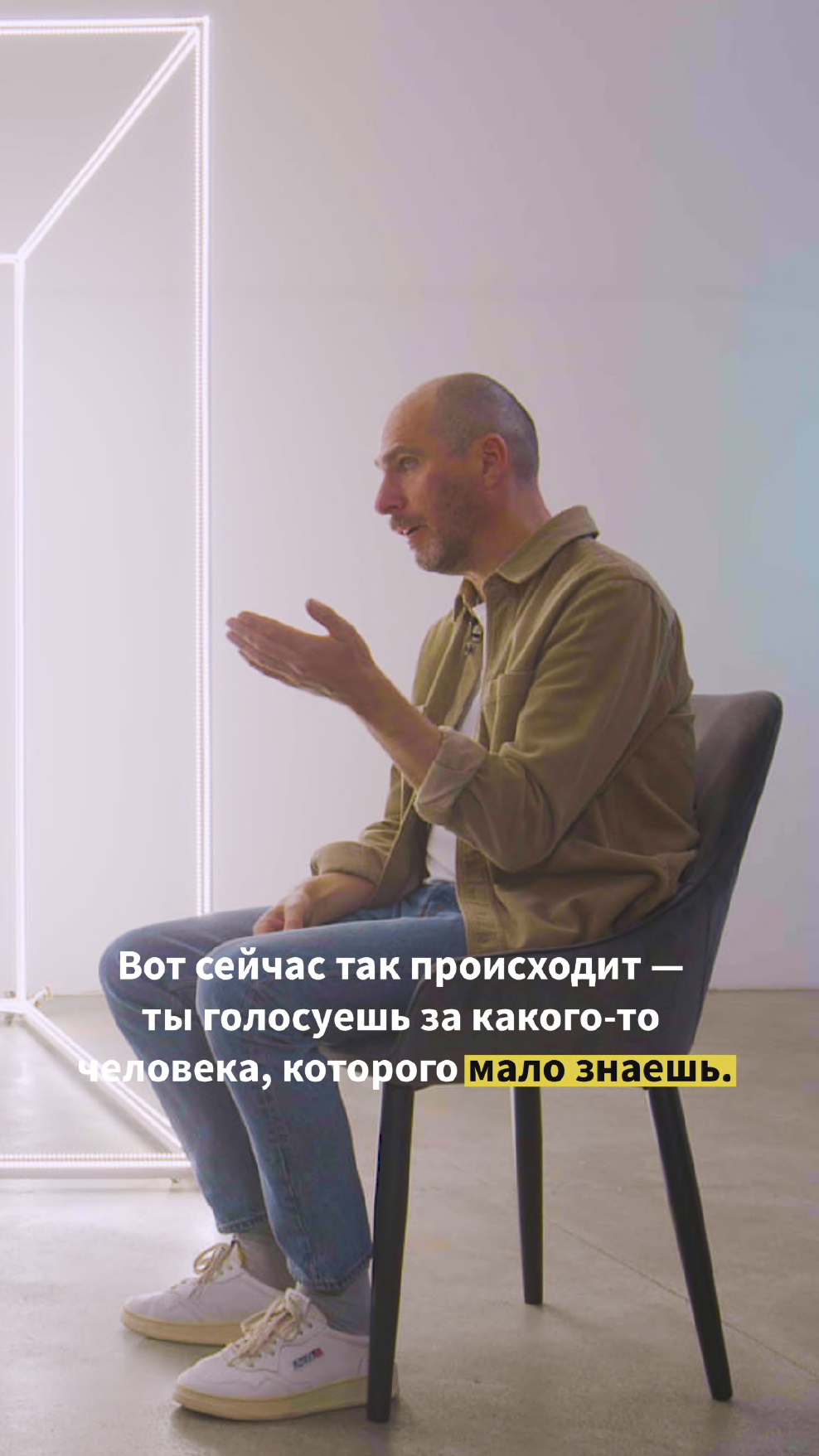}
    \includegraphics[width=0.48\linewidth, trim={0cm 0cm 0cm 0cm}, clip]{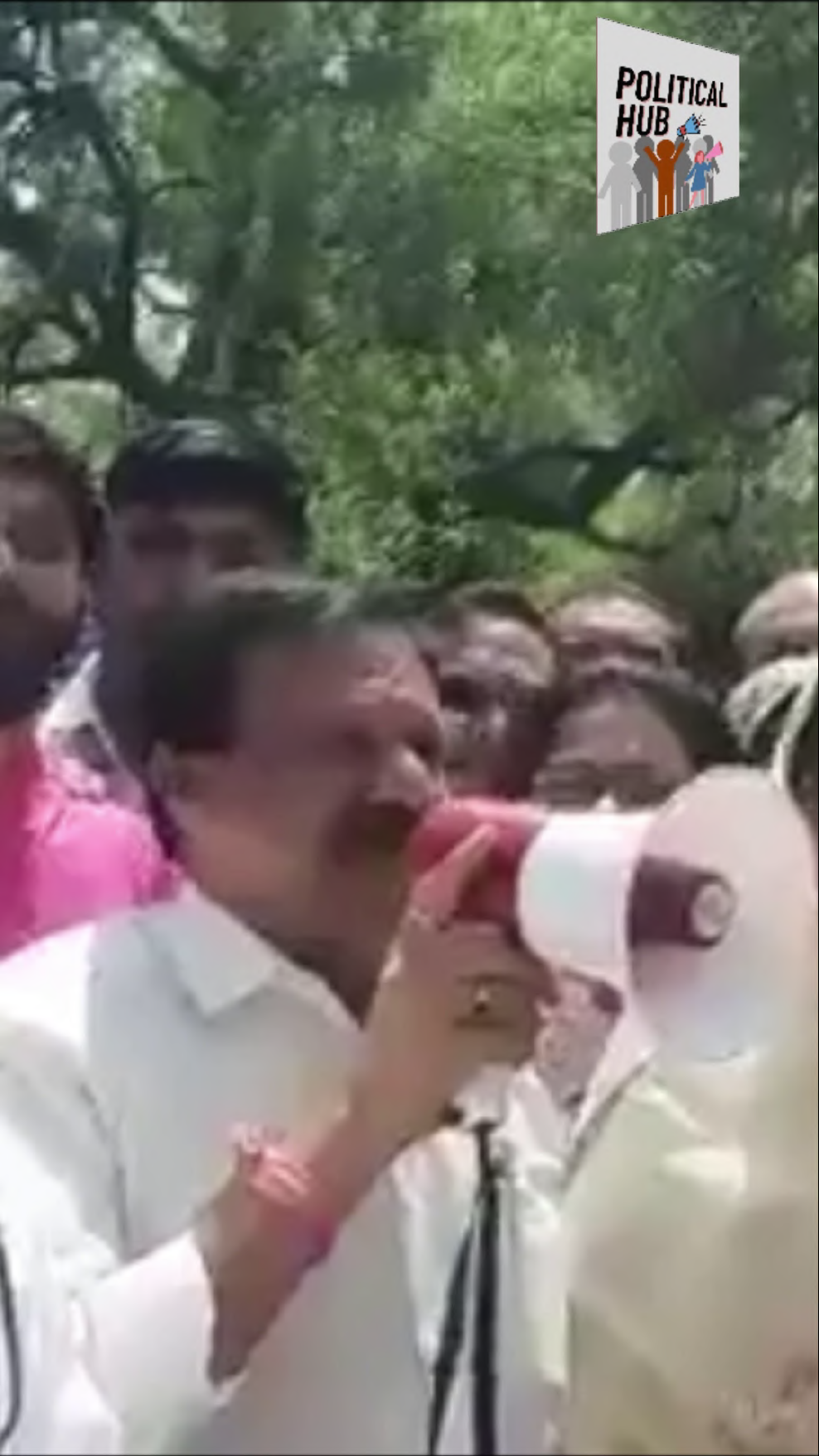}
    \captionsetup{labelformat=empty}
    \caption{Society}
  \end{subfigure}
  \hspace{0.01 in}
  \begin{subfigure}{0.18\linewidth}
    \centering
    \includegraphics[width=0.48\linewidth, trim={0cm 0cm 0cm 0cm}, clip]{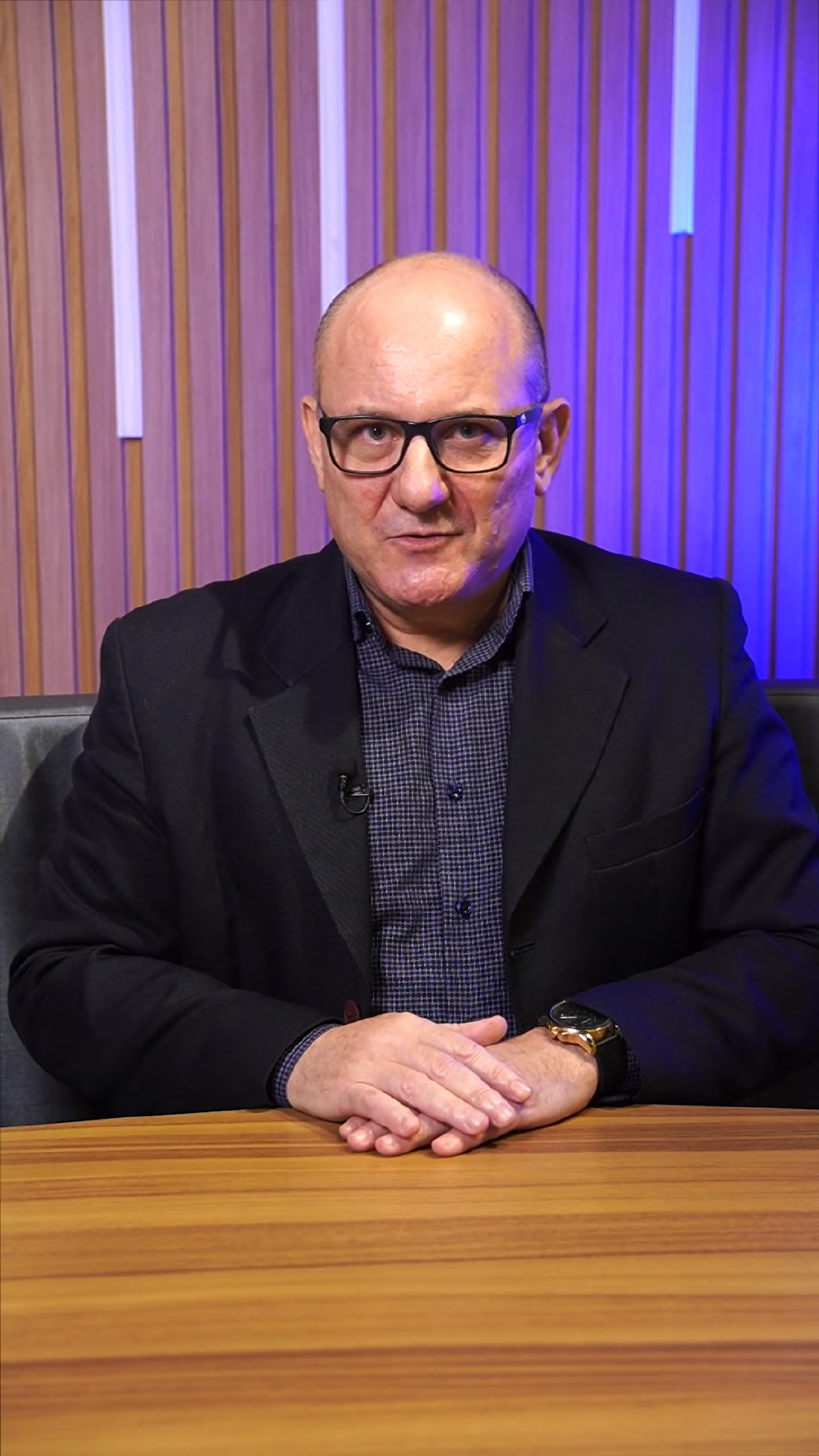}
    \includegraphics[width=0.48\linewidth, trim={0cm 0cm 0cm 0cm}, clip]{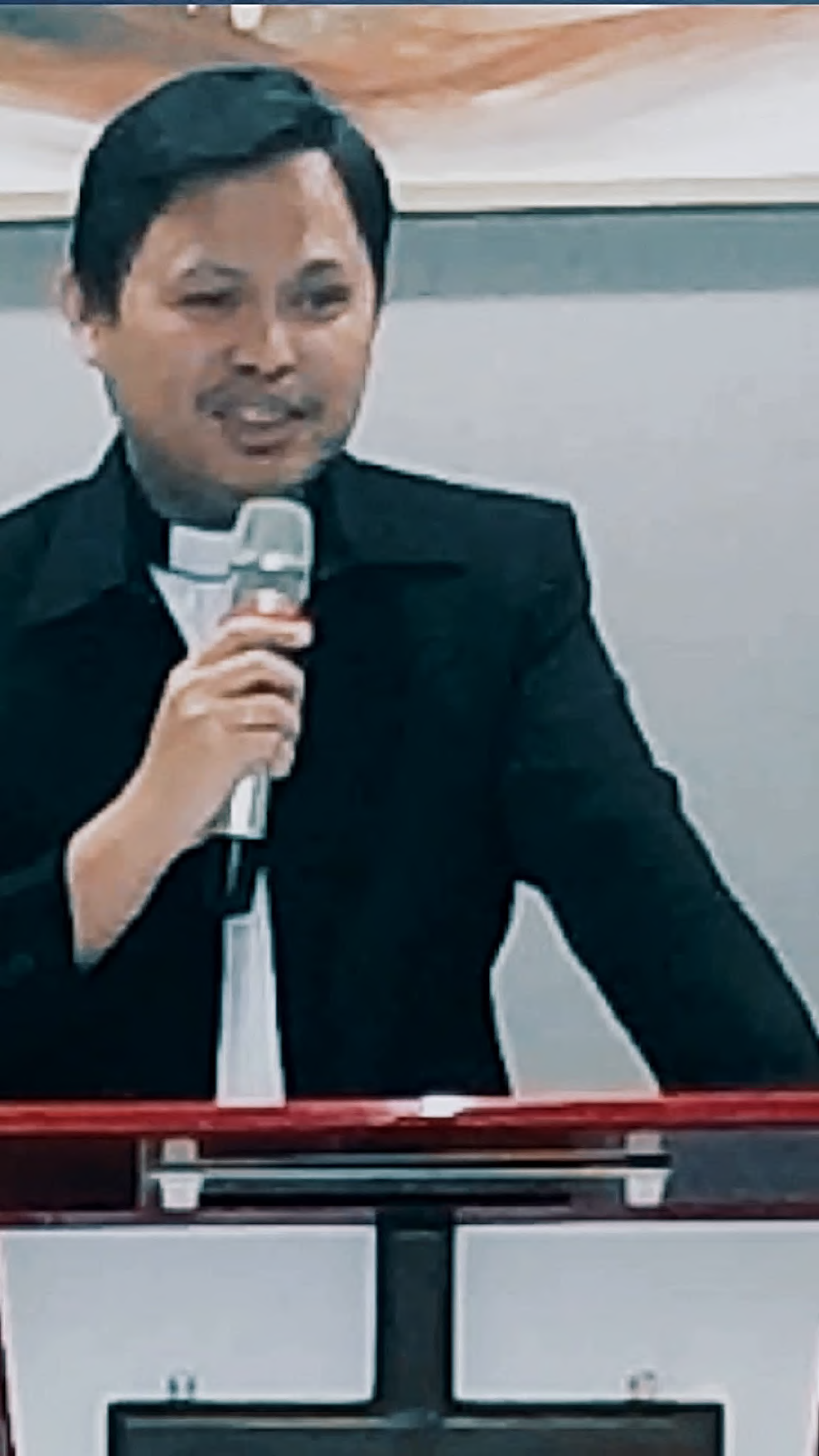}
    \captionsetup{labelformat=empty}
    \caption{Speech}    
  \end{subfigure}
  \hspace{0.01 in}
  \begin{subfigure}{0.18\linewidth}
    \centering
    \includegraphics[width=0.48\linewidth, trim={0cm 0cm 0cm 0cm}, clip]{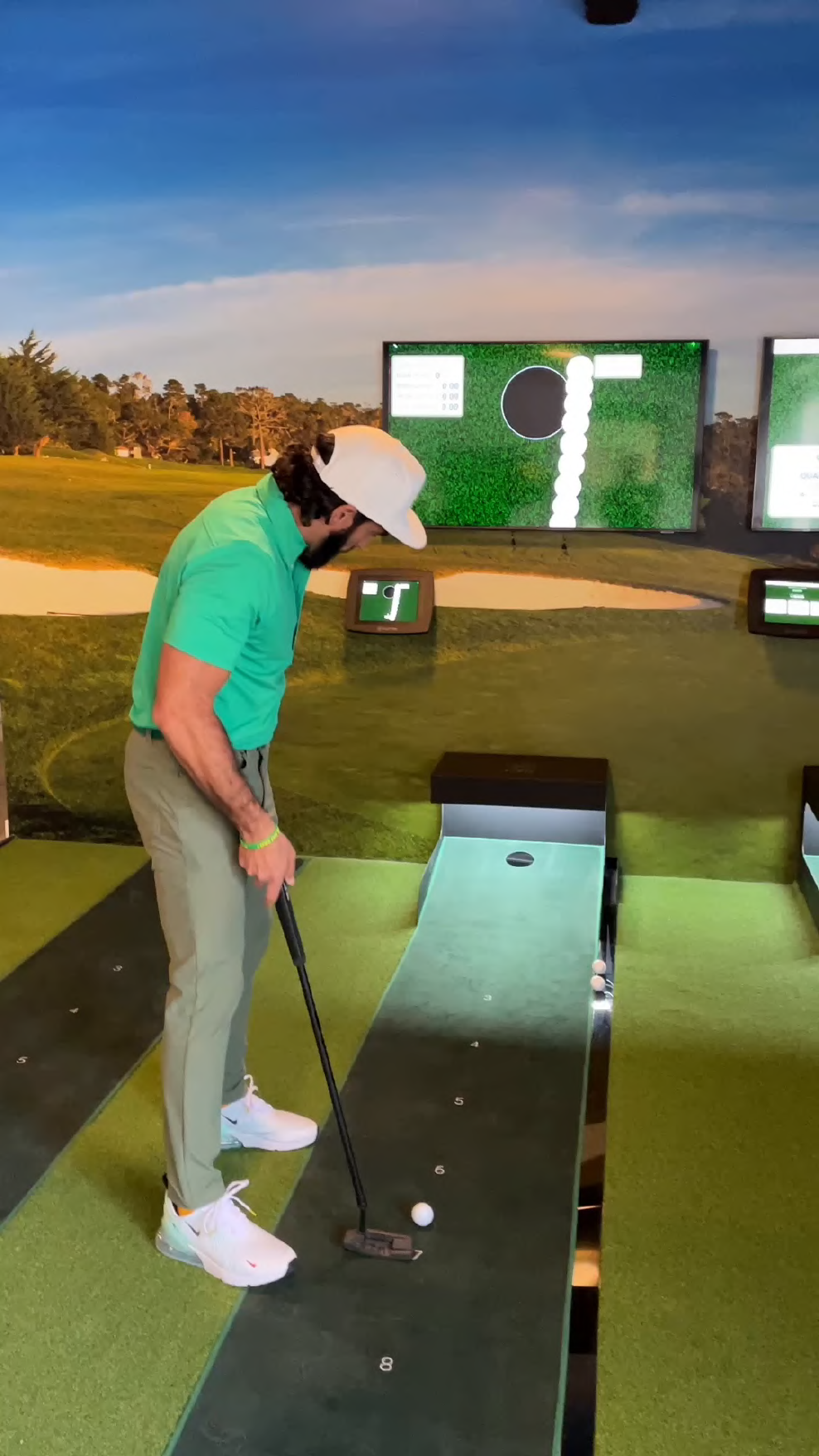}
    \includegraphics[width=0.48\linewidth, trim={0cm 0cm 0cm 0cm}, clip]{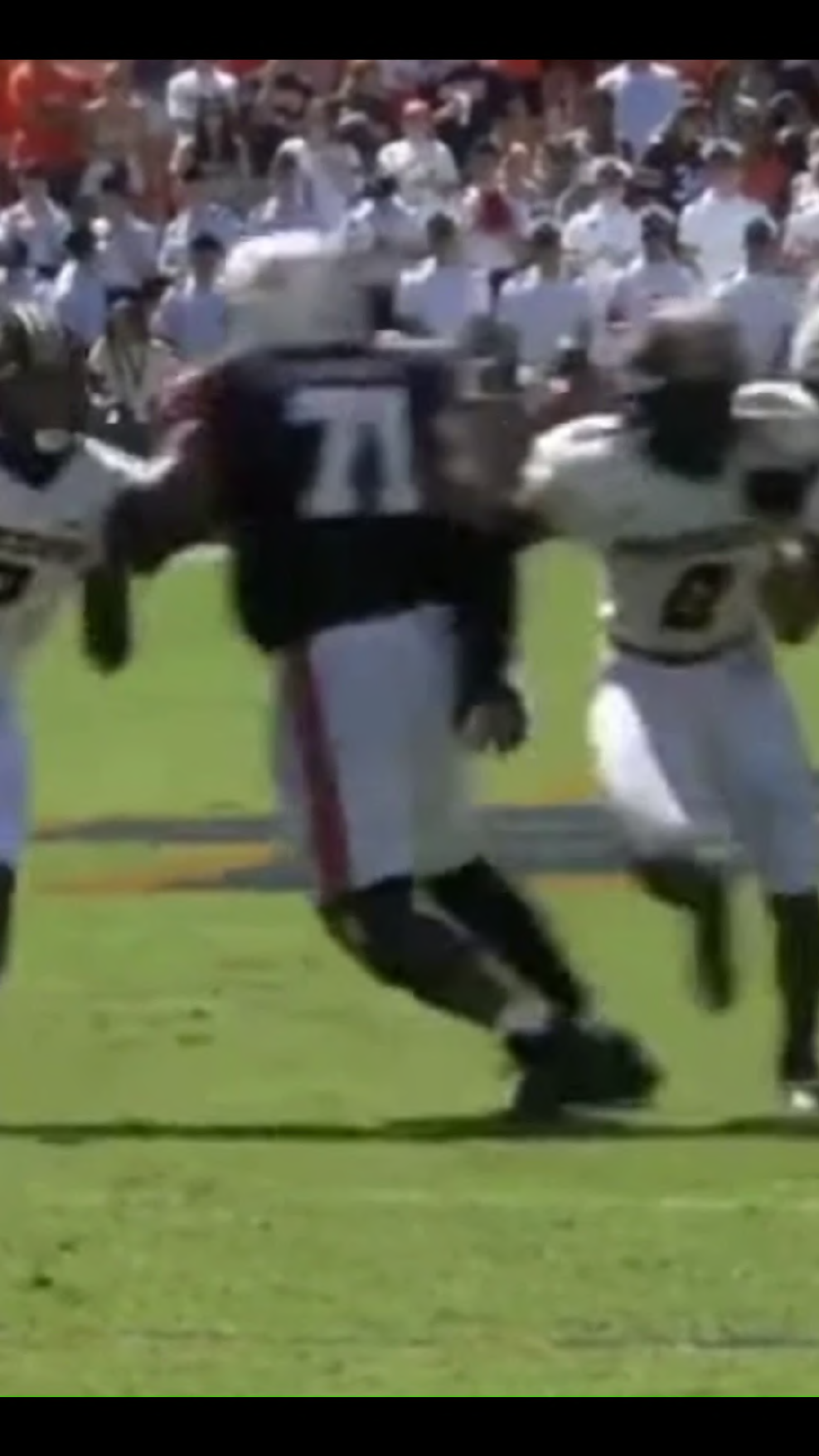}
    \captionsetup{labelformat=empty}
    \caption{Sports}    
  \end{subfigure}\\
\caption{Samples in high (left) and low (right) quality per category.}
\label{fig:samples_high_low_uvq}
\end{figure*}

Figure~\ref{fig:distribution_sitiuvq} shows the scatter plots for pairs of SI, TI, and UVQ for three content categories: Cooking, Health, and Gameplay. We can see most Cooking videos have low SI and high TI, while most Health videos have relatively low TI. Both Cooking and Health videos have relatively high quality (UVQ $>4.0$), while Gameplay videos have wider distribution in all three dimensions.
This analysis demonstrated that the content categories had significant discrepancy, and needed to be investigated separately if possible. We used the mean values of SI, TI, and UVQ to divide the entire pool into 8 ($=2\times{2}\times{2}$) subregions, and randomly selected equal number of samples from each subregion for each content category. This sampling strategy includes more samples from low density regions, which effectively suppressed the bias of original data distribution. In practice, we selected 50 samples per subregion per category for SDR SFV. We keep the entire HDR set for the final review since its total size is already close to the target size (4000 v.s. 2000). After this step, both SDR and HDR pools remain about 4000 samples. Fig.~\ref{fig:samples_high_low_uvq} shows samples in high and low quality (approximated by UVQ) for each content category, where we can see a distinct gap between high and low quality samples. Similar distinct gaps can be found in SI and TI, which demonstrated good diversity.

\subsection{Final Content Review}
\label{subsec:Final Content Review}
\begin{figure}
\centering
\begin{subfigure}{1.0\linewidth}
    \centering
    \includegraphics[width=0.19\textwidth, trim={0cm 0cm 0cm 0cm}, clip]{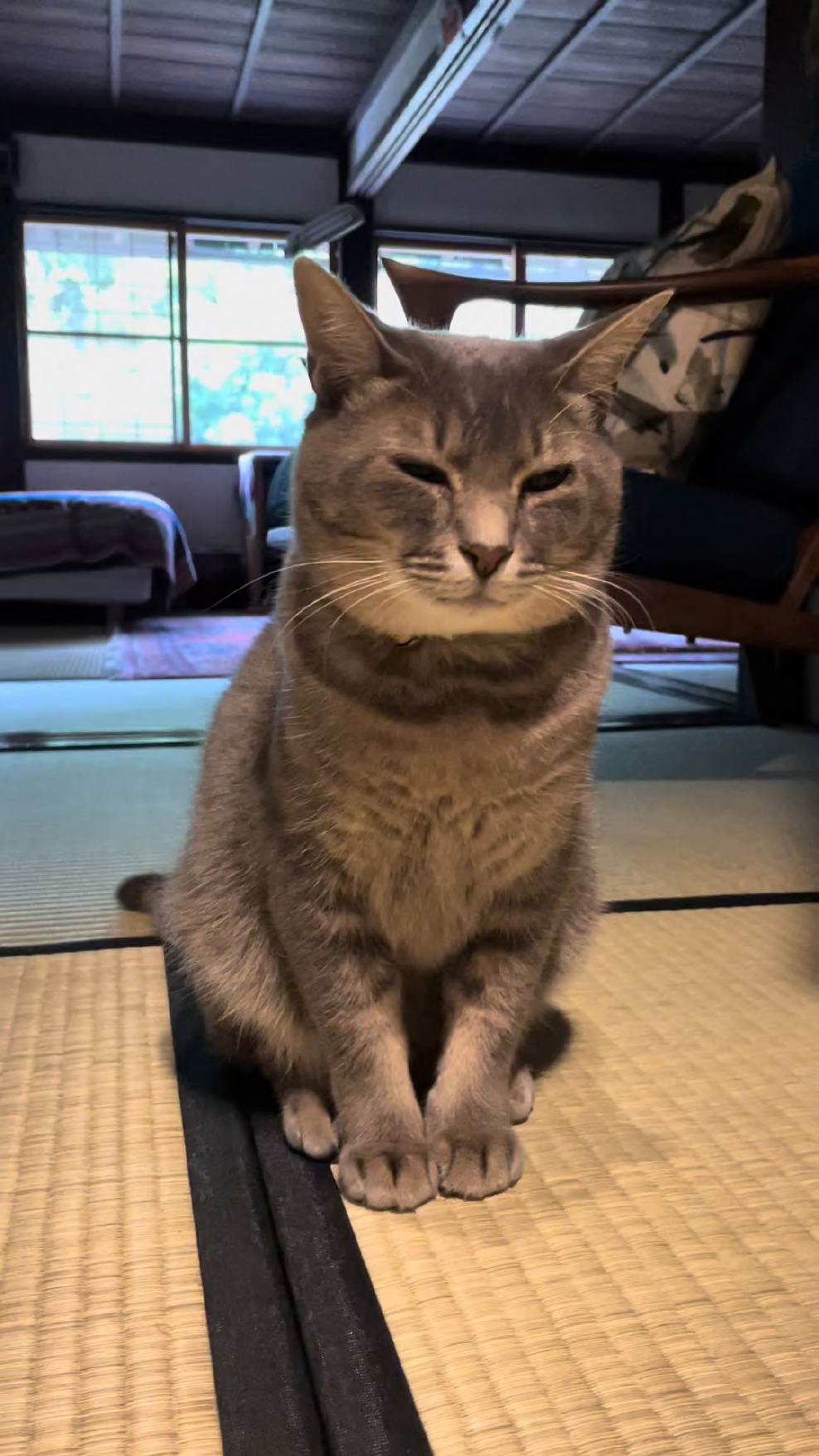}
    \includegraphics[width=0.19\textwidth, trim={0cm 0cm 0cm 0cm}, clip]{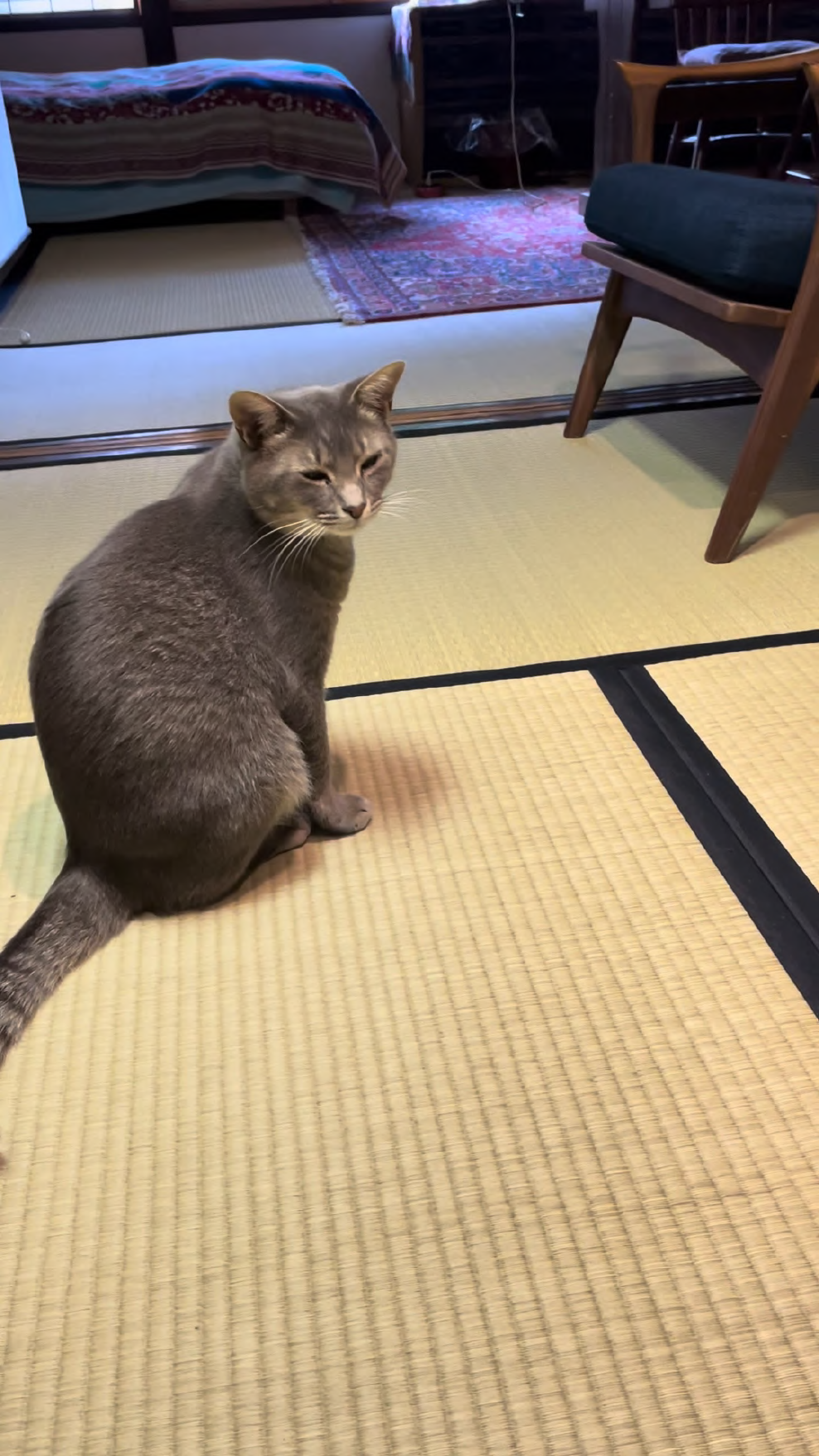}
    \includegraphics[width=0.19\textwidth, trim={0cm 0cm 0cm 0cm}, clip]{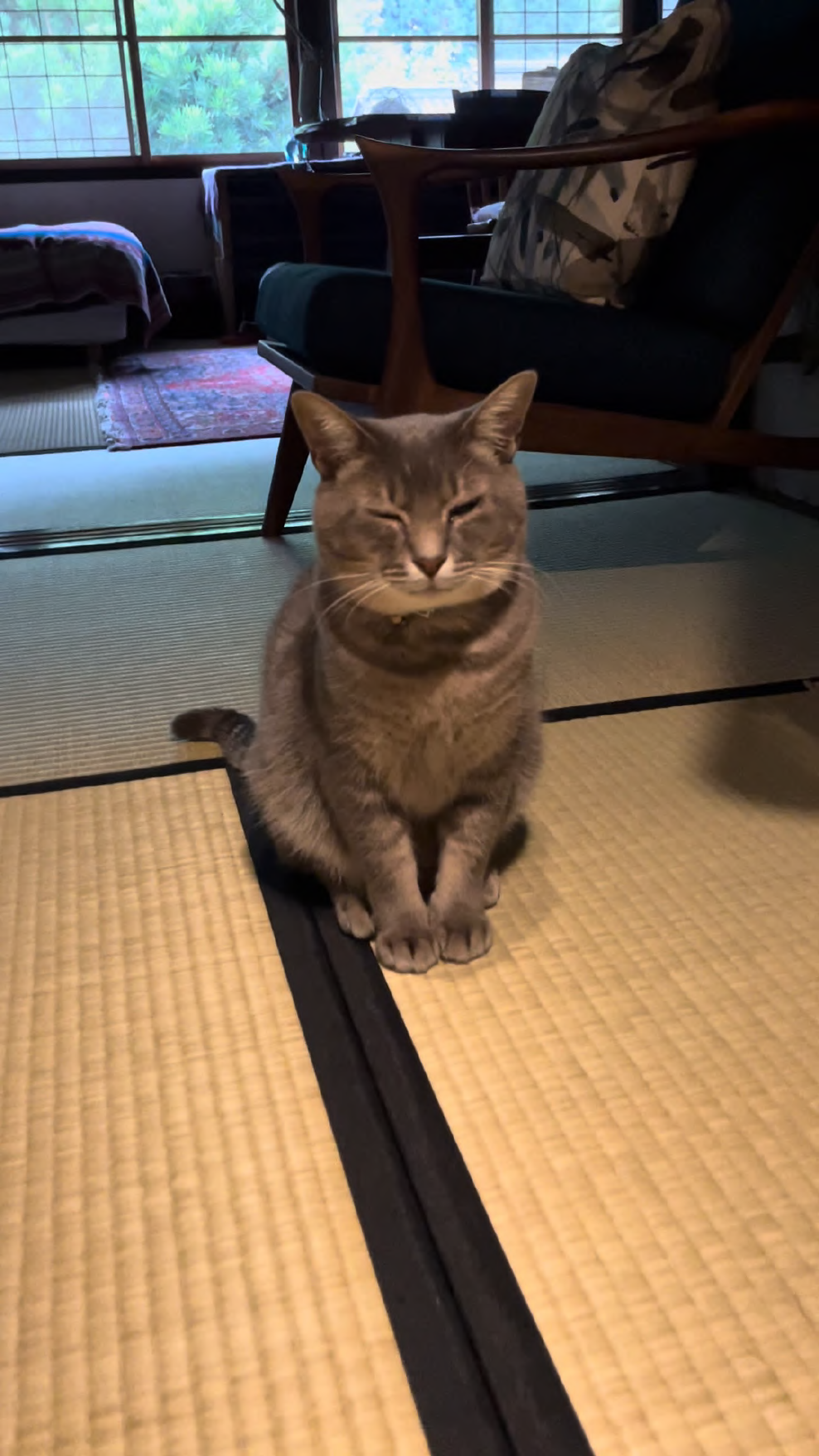}
    \includegraphics[width=0.19\textwidth, trim={0cm 0cm 0cm 0cm}, clip]{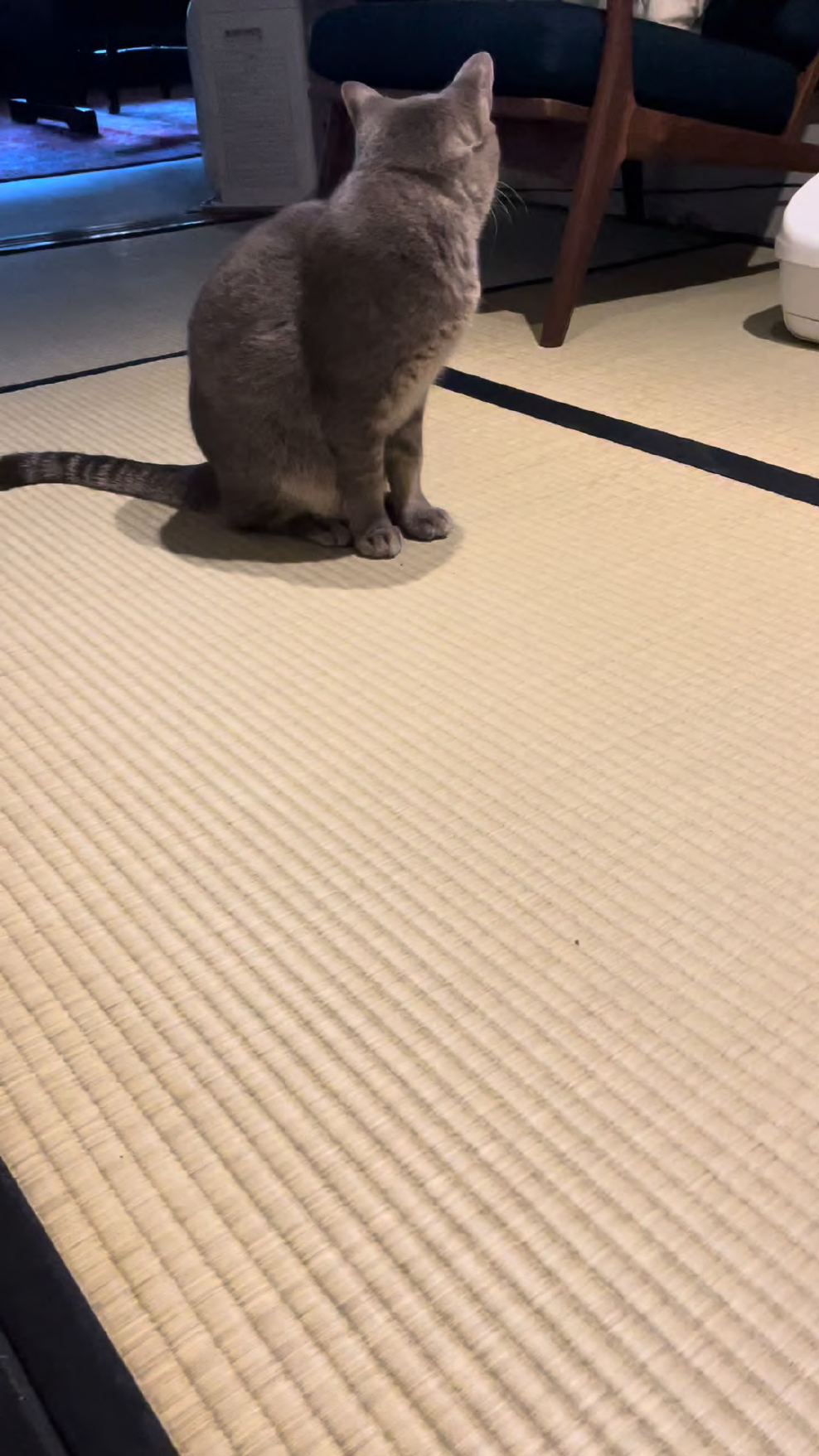}
    \includegraphics[width=0.19\textwidth, trim={0cm 0cm 0cm 0cm}, clip]{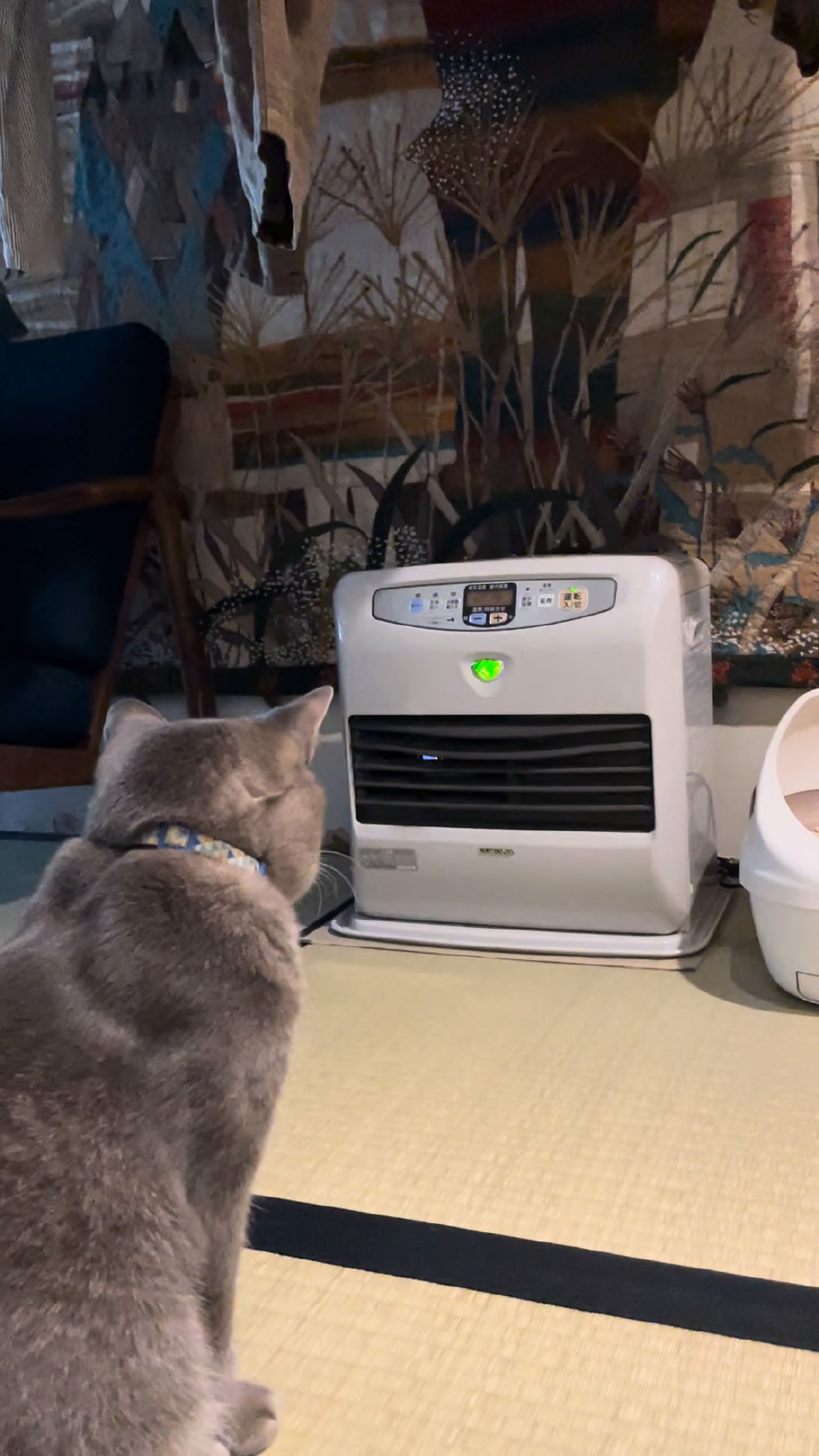}
    \caption{Random sampling (many similar contents.)}
\end{subfigure}\\
\begin{subfigure}{1.0\linewidth}
    \centering
    \includegraphics[width=0.19\textwidth, trim={0cm 0cm 0cm 0cm}, clip]{figures/HDR_Animal_kopm_0001-min.pdf}
    \includegraphics[width=0.19\textwidth, trim={0cm 0cm 0cm 0cm}, clip]{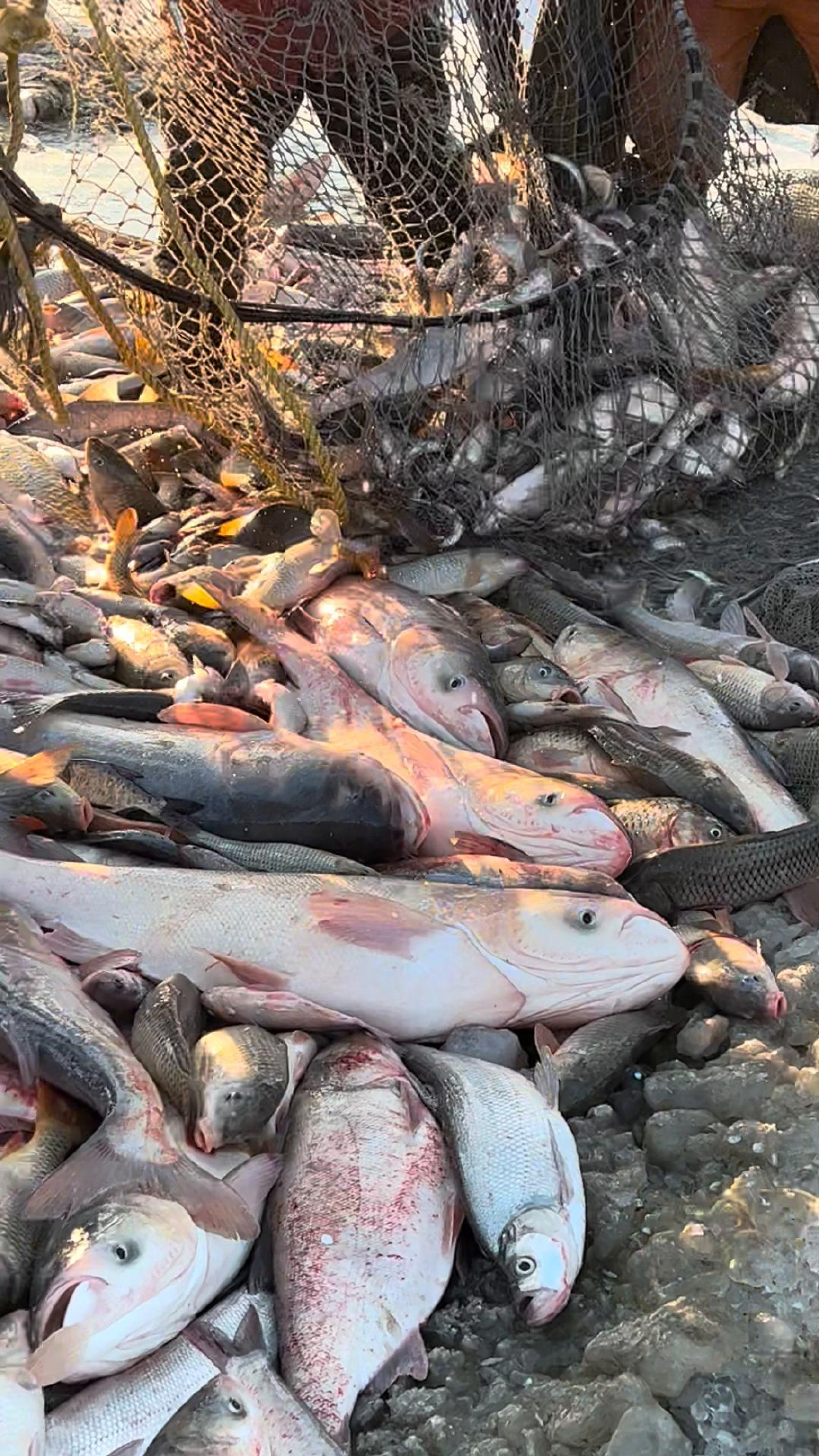}
    \includegraphics[width=0.19\textwidth, trim={0cm 0cm 0cm 0cm}, clip]{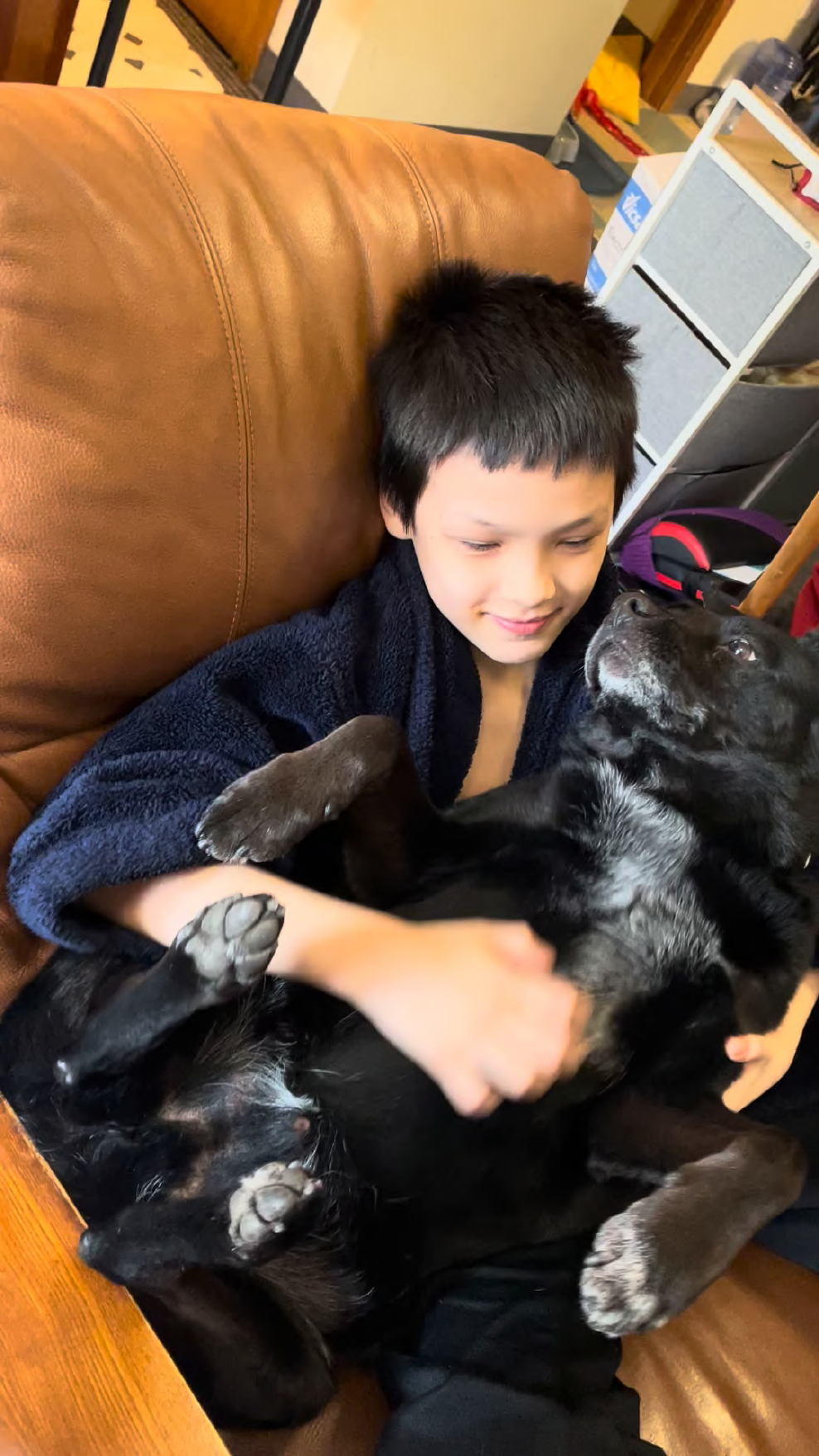}
    \includegraphics[width=0.19\textwidth, trim={0cm 0cm 0cm 0cm}, clip]{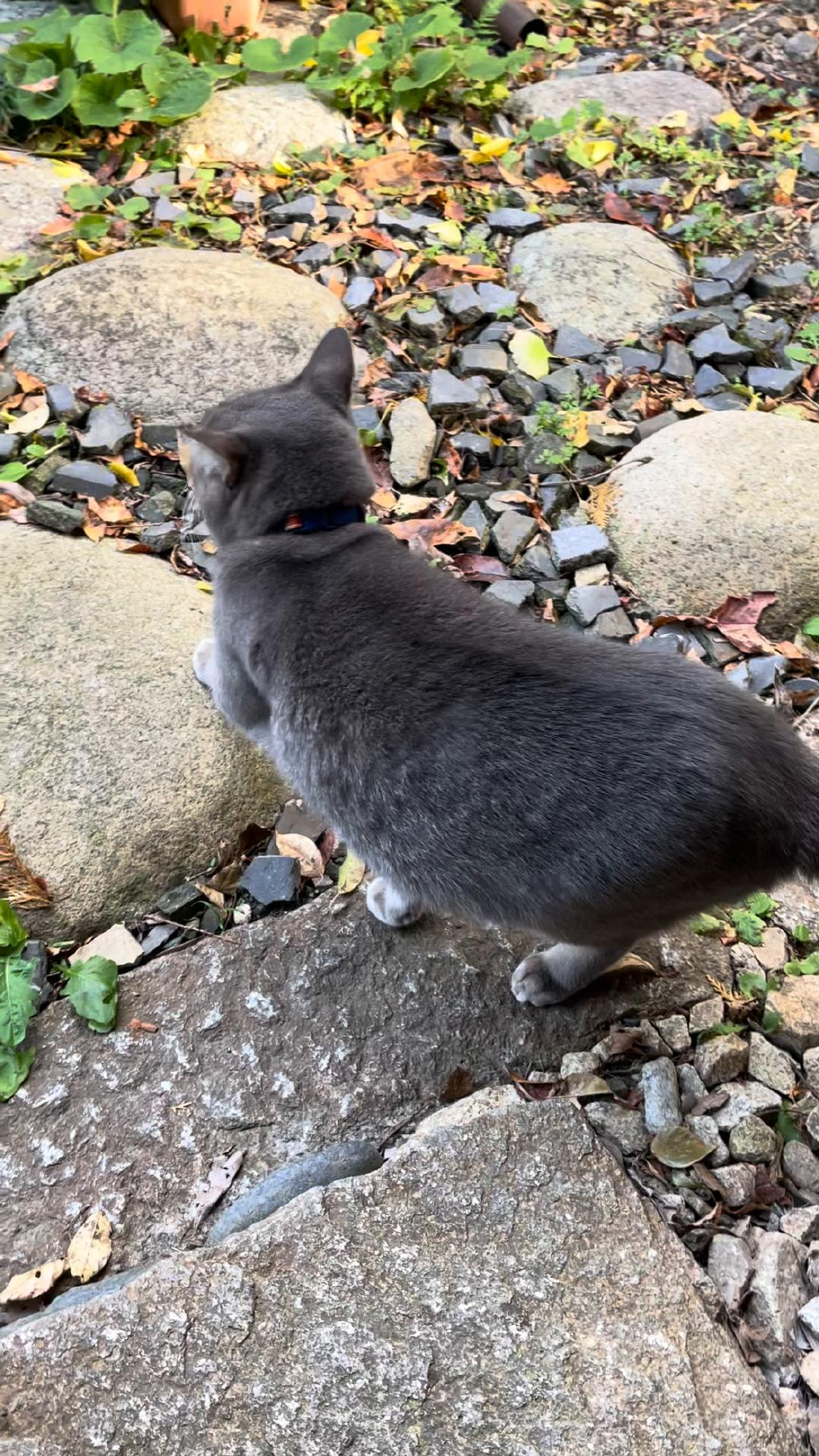}
    \includegraphics[width=0.19\textwidth, trim={0cm 0cm 0cm 0cm}, clip]{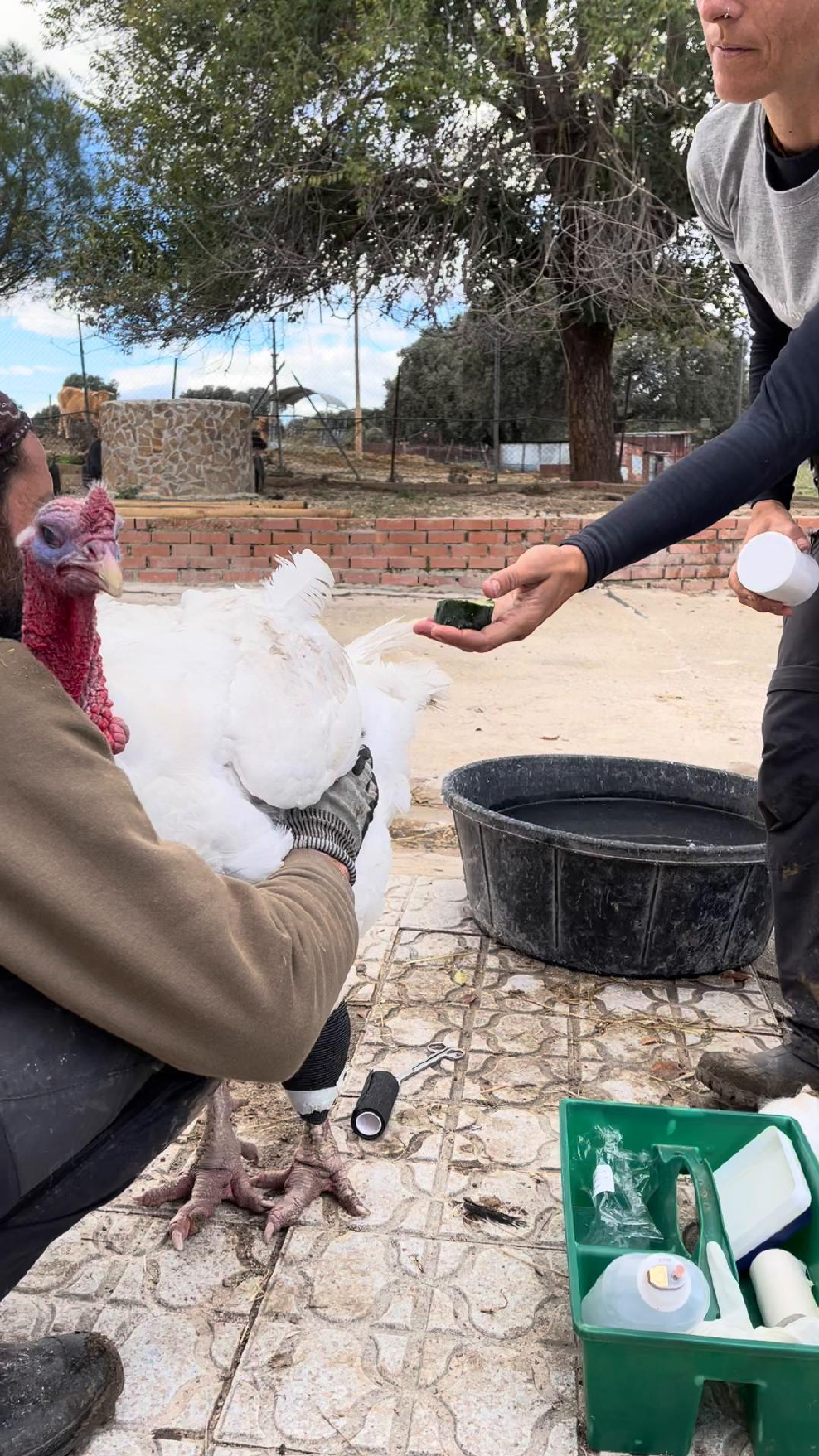}
    \caption{Manual sampling (better content diversity)}
\end{subfigure}\\
\caption{Random sampling v.s. manual sampling.}
\label{fig:manual_sampling}
\end{figure}

Step 1 and 2 in our sampling framework mainly rely on objective features (e.g. content category) and metrics (SI, TI, and UVQ), which still has some problems. For example, videos with inappropriate contents cannot be filtered out by above objective metrics. Also to maximize the diversity of the dataset, it is preferable to remove duplicates and limit the number of similar samples. 
This duplicates issue is more severe in our HDR pool than the SDR pool, due to limited candidates. 
Several creators contributed hundreds of samples with similar contents, which could lead to a significant bias of the final dataset.
Thus a careful manual review is a necessary step to finalize the dataset.
For this SFV+HDR dataset, we cleaned up the content with multiple manual reviews to maximize the content diversity, as shown in Fig.~\ref{fig:manual_sampling}. 

After this three step sampling, we finally selected 2030 SDR and 2000 HDR samples. For SDR, most content categories have 210 samples except Dance (140 samples). For HDR, we selected 30 samples per content category, and the other 1700 samples with unknown category.

%% file: subjective_data_analysis.tex
\section{Subjective Data Analysis}
\label{sec:Subjective Data Analysis}
\subsection{Subjective Experiment}
\label{subsec:Subjective Experiment}
With the finalized video set, the next challenge is to collect subjective quality scores. Original YouTube uploads are in various formats (e.g., different codecs, color spaces, and frame rates). In order to be playable on all clients' browsers, we transcoded all original videos using H.264~\cite{ieee2003h264} with YUV420P and Constant Rate Factor (CRF) value of 10 to preserve the quality as close as possible to the original version. For HDR videos, two major types are Perceptual Quantization (PQ) and Hybrid Log-Gamma (HLG). We converted these HDR videos to SDR using corresponding tone mappings. 
The subjective tests were run on mobile phone, since it is the main device interface for SFV. 

We conducted two subjective tests for SDR and HDR respectively. The SDR test contained 4030 (2030 native SDR and 2000 HDR2SDR) videos. 66 subjects (from an internal data labeling team) participated in this SDR test using their own mobile phones. Each subject participated in 7 sessions, where each session contained 300 randomly selected SDR videos. 
The HDR test included 300 native HDR videos and 300 corresponding HDR2SDR versions. After preliminary tests, we found HDR experiences were highly different on different devices. For example, the same HDR videos look much brighter on Pixel 7 pro than on Pixel 5, due to different peak brightness. To get more reliable subjective data, 10 subjects who used the same phone (Pixel 7 pro) participated in an additional HDR test (2 sessions, 300 videos per session). 
Subjects were first shown three training videos to get them familiar with the testing process. These three videos were chosen to exemplify bad, okay, and good qualities. After the training, subjects were presented testing videos that were randomly sampled from the entire dataset. For each video, subjects had to watch the entire duration of the clip, and they were allowed to replay the video if necessary. Then they were asked the quality assessment question of \textit{How was the overall video quality?} The rating was given on a 1 to 5 scale slider, adjustable in 0.1 increments, where each integer is marked as \textit{Bad (1)}, \textit{Poor (2)}, \textit{Fair (3)}, \textit{Good (4)}, and \textit{Excellent (5)}.
The test was designed to be finished within 30 minutes. Each SDR video clip was finally rated by 25 to 40 subjects, and HDR videos had 10 ratings. Since our raters were from a professional data labeling team, whose ratings were reliable, we used all their ratings to compute Mean Opinion Score (MOS).

\subsection{SDR MOS Analysis}
\label{subsec:SDR MOS Analysis}
\begin{figure}
\centering
\includegraphics[width=0.31\linewidth, trim={0cm 0cm 0cm 0cm}, clip]{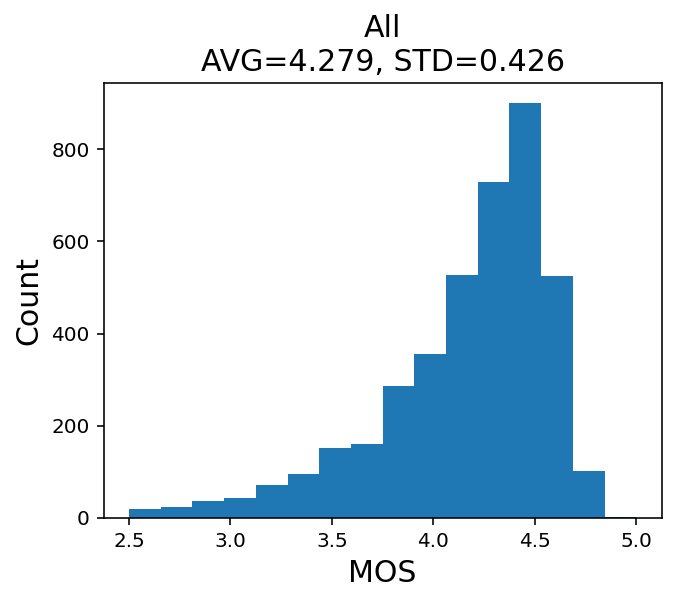}
\includegraphics[width=0.31\linewidth, trim={0cm 0cm 0cm 0cm}, clip]{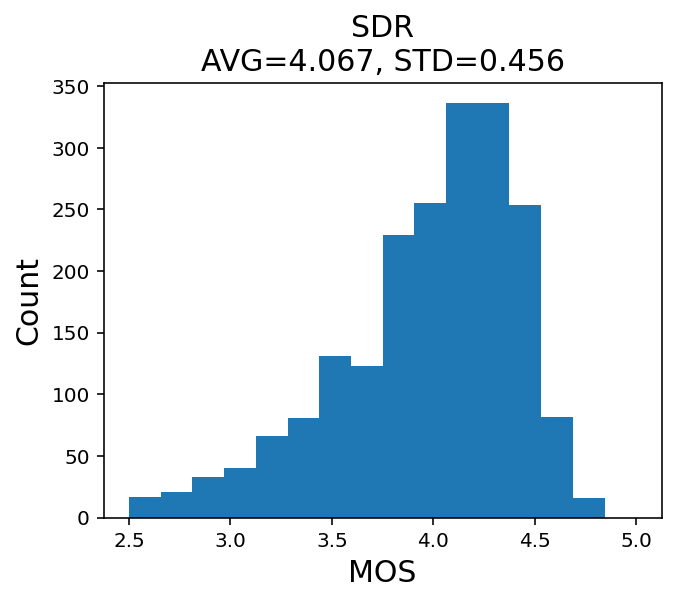}
\includegraphics[width=0.31\linewidth, trim={0cm 0cm 0cm 0cm}, clip]{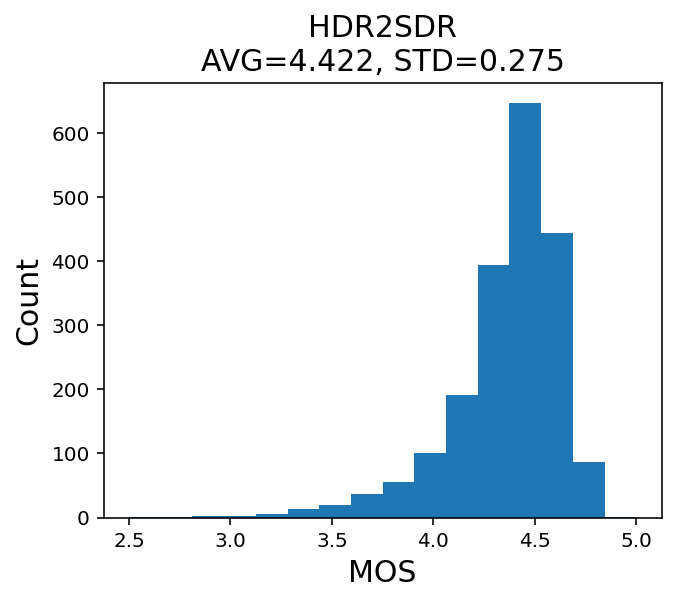}
\caption{MOS distributions of all SDR videos, native SDR, and HDR2SDR respectively.}
\label{fig:mos_voerall}
\end{figure}

\begin{figure}
\centering
\begin{subfigure}{0.35\linewidth}
\centering
\includegraphics[width=1\linewidth, trim={0cm 0cm 0cm 0cm}, clip]{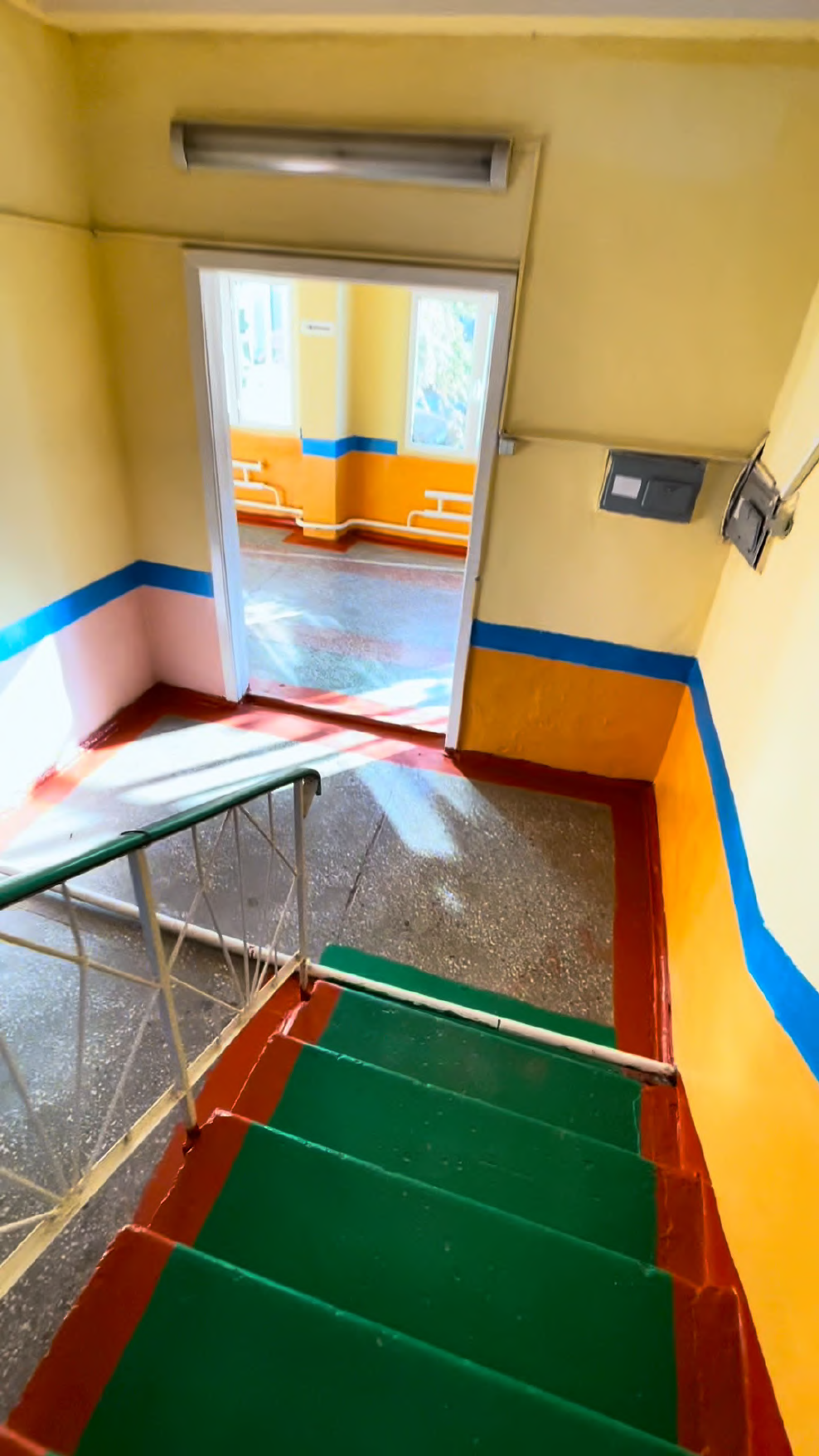}
\captionsetup{labelformat=empty, justification=centering}
\caption{HDR\_Unknown\_2teo (MOS=4.27)}
\end{subfigure}
\begin{subfigure}{0.35\linewidth}
\centering
\includegraphics[width=1\linewidth, trim={0cm 0cm 0cm 0cm}, clip]{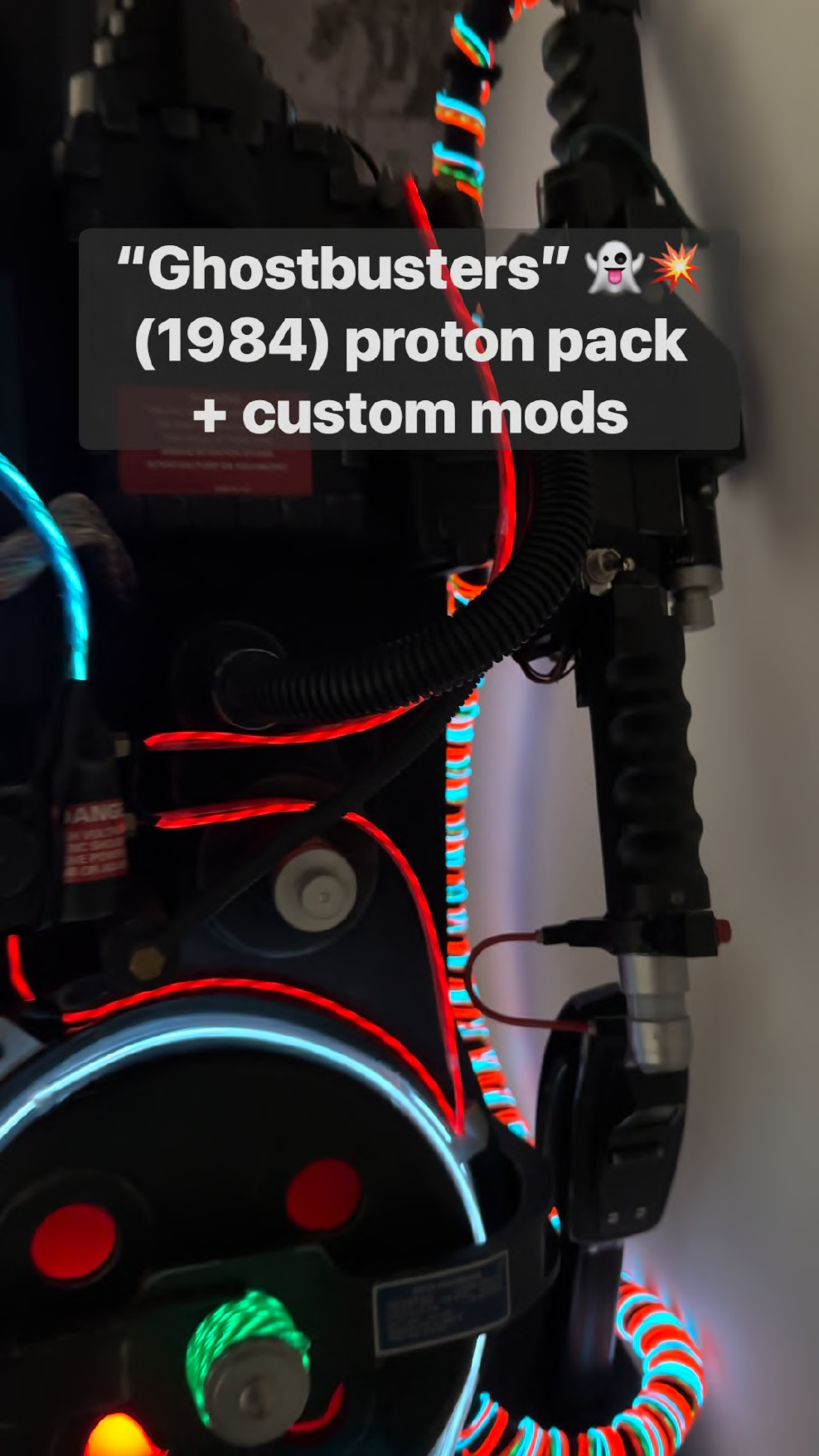}
\captionsetup{labelformat=empty, justification=centering}
\caption{HDR\_Unknown\_faf2 (MOS=4.29)}
\end{subfigure}
\caption{HDR2SDR samples with high MOS, even though there are some noticeable artifacts.}
\label{fig:hdr_samples}
\end{figure}

The left histogram in Fig.~\ref{fig:mos_voerall} showed the overall MOS distribution for the entire SDR set (4030 samples, including native SDR and HDR2SDR). We found that the MOS distribution is relatively narrow, where $80\%$ MOS values are within [3.8, 4.6]. It suggested that many SFV have equally good quality, and meet viewers' quality expectation.
We further broke down the set into native SDR and HDR2SDR (middle and right histograms in Fig.~\ref{fig:mos_voerall}). We can clearly see that MOS of most HDR2SDR ($90\%$) are higher than 4.0, which has two potential reasons: 1) HDR videos are usually captured by high end devices which natively provides high picture quality; 2) The color plays an important role in quality assessment. The first reason is intuitively major, but we also found some examples supporting the second reason. As shown in Fig.~\ref{fig:hdr_samples}, the two HDR2SDR samples have saturated color and keep more local color details, which makes them look professional and high quality. It implies that good attributes of HDR are still maintained in their HDR2SDR versions.

Content dependency is also an important feature of UGC quality assessment. Fig.~\ref{fig:mos_sdr_per_category} shows the MOS distributions of native SDR for individual categories. We can see that Society and Speech have relatively uniform distributions (and lower average quality) than other categories. One potential reason is that many content were recorded in public spaces with restricted lighting and device control. Another potential indication is that viewers are not very interested in such contents and 
intuitively avoid giving very high scores. In contrast, Cooking and Hobby have the highest average quality, since the contents are widely interesting, and creators fully control the recording environments and are able to do sophisticated post-enhancements. Above are just preliminary observations. Understanding the content impact on perceptual quality is an important topic with practical impacts (e.g. calibrating quality score among different contents). We hope our dataset encourage more thorough research on this topic.

\subsection{HDR MOS Analysis}
\label{subsec:HDR MOS Analysis}
\begin{figure}
\centering
\includegraphics[width=1.0\linewidth, trim={0cm 0cm 0cm 0cm}, clip]{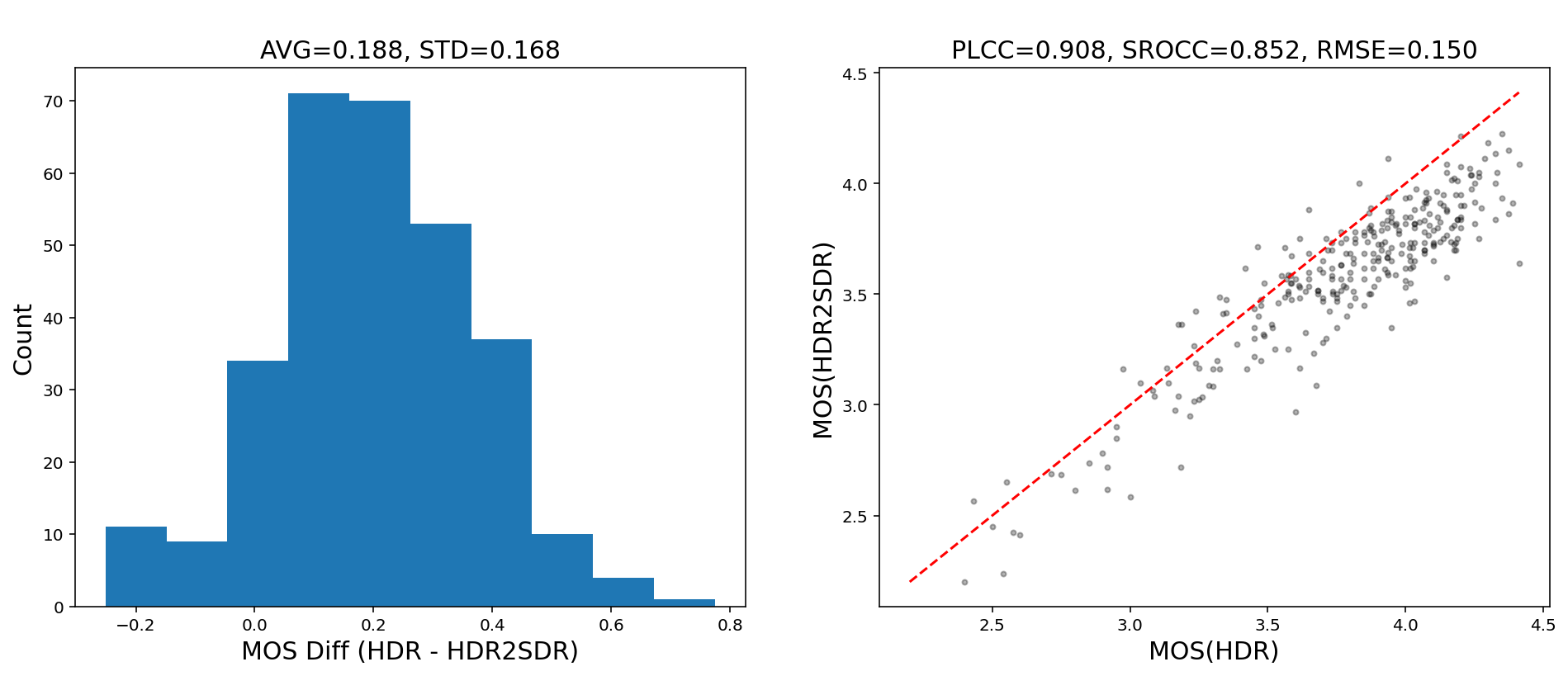}
\caption{MOS comparison between HDR and HDR2SDR, where most native HDR videos have higher MOS than corresponding HDR2SDR versions.}
\label{fig:hdr_vs_hdr2sdr}
\end{figure}

Fig.~\ref{fig:hdr_vs_hdr2sdr} compared MOS for native HDR and corresponding HDR2SDR versions.  We can see most HDR MOS are higher than corresponding HDR2SDR version, and the average MOS difference is about 0.18, which means a significant gap. Based on raters' feedback, the brightness played an important role in the quality assessment. HDR videos are significantly brighter with more clear details than SDR versions, which makes HDR look better.

% \begin{figure*} 
% \centering
% \includegraphics[width=1\linewidth, trim={0cm 0cm 0cm 0cm}, clip]{figures/mos_sdr_per_category.png}
% \caption{MOS distributions per content category for native SDR videos.}
% \label{fig:mos_sdr_per_category}
% \end{figure*}

%% file: objective_metric_performance.tex
\section{Objective Metric Performance}
\label{sec:Objective Metric Performance}
UGC quality has been studied for years, and SOTA no reference metrics have achieved good correlation with subjective scores. It is interesting to see how well they perform on SFV contents. Table~\ref{tab:mos_correlations} shows the MOS correlations with three SOTA UGC metrics: DOVER~\cite{Wu2023DOVER}, FAST-VQA~\cite{Wu2022FASTVQA}, and FasterVQA~\cite{Wu2022FasterVQA}. All metric scores are rescaled to the MOS range ([1, 5]). We can see FAST-VQA has the highest overall MOS correlations (0.79), DOVER and FasterVQA's PLCCs are also above 0.75. It implies that SFV quality assessment is not a brand new problem, and SOTA VQA models can be reused to achieve reasonable accuracy. 
We observed that these metrics have worse correlations on HDR2SDR videos (best PLCC is 0.66). However, it is insufficient to conclude that HDR2SDR quality assessment is more difficult than SDR assessment, because these two types of videos have different MOS distributions. $90.4\% (=1808)$ HDR2SDR videos are relatively high quality (MOS $\ge{4.0}$), while only $56.2\%(=1141)$ native SDR videos are above 4.0. To fairly evaluate their difference, we randomly selected 500 videos from both high quality sets and computed corresponding correlations. This process was repeated for 1000 times, and their average correlations were reported in Table~\ref{tab:high_mos_correlations}. We can see PLCCs on native SDR are still significantly (0.05 to 0.07) higher than HDR2SDR versions, which more or less demonstrates the difficulty of HDR2SDR videos, and implies that color sensitivity is a potential topic for future VQA research.

\begin{table}
\footnotesize
\centering
\begin{tabular}{l c c c c c c}
\toprule
  & \multicolumn{2}{c}{All} & \multicolumn{2}{c}{Native SDR} & \multicolumn{2}{c}{HDR2SDR} \\
  \cmidrule(r){2-3} \cmidrule{4-5} \cmidrule(l){6-7}
          & PLCC  & SRCC  & PLCC  & SRCC  & PLCC  & SRCC  \\ \hline
DOVER     & 0.781 & 0.702 & 0.793 & 0.750 & 0.618 & 0.496  \\ 
FAST-VQA  & 0.797 & 0.752 & 0.789 & 0.789 & 0.664 & 0.543  \\ 
FasterVQA & 0.755 & 0.705 & 0.753 & 0.748 & 0.585 & 0.493  \\ 
\bottomrule
\end{tabular}
\caption{MOS correlations for all, native SDR, and HDR2SDR videos.}
\label{tab:mos_correlations}
\end{table}

\begin{table}
\footnotesize
\centering
\begin{tabular}{l c c c c}
\toprule
  & \multicolumn{2}{c}{Native SDR} & \multicolumn{2}{c}{HDR2SDR} \\
  \cmidrule(r){2-3} \cmidrule(l){4-5}
          & PLCC  & SRCC  & PLCC  & SRCC  \\ \hline
DOVER     & 0.468 & 0.486 & 0.414 & 0.379 \\ 
FAST-VQA  & 0.497 & 0.539 & 0.429 & 0.422 \\ 
FasterVQA & 0.423 & 0.485 & 0.353 & 0.378 \\ 
\bottomrule
\end{tabular}
\caption{High range MOS correlations for native SDR and HDR2SDR videos.}
\label{tab:high_mos_correlations}
\end{table}

\begin{table}
\footnotesize
\centering
\begin{tabular}{l c c c}
\toprule
Category  & DOVER        & FAST-VQA    & FasterVQA \\
          & PLCC/SRCC    & PLCC/SRCC   & PLCC/SRCC\\ \hline
Animal    &  0.848/0.775 & 0.829/0.793 & 0.786/0.735 \\ \hline
Cooking   &  0.753/0.731 & 0.733/0.775 & 0.646/0.664  \\ \hline
Dance     &  0.883/0.851 & 0.882/0.866 & 0.860/0.833 \\ \hline
Gameplay  &  0.639/0.545 & 0.634/0.557 & 0.640/0.558 \\ \hline
Health    &  0.784/0.691 & 0.810/0.768 & 0.745/0.712 \\ \hline
Hobby     &  0.596/0.568 & 0.708/0.693 & 0.606/0.617 \\ \hline
Music     &  0.772/0.724 & 0.738/0.721 & 0.745/0.728 \\ \hline
Society   &  0.842/0.843 & 0.770/0.796 & 0.759/0.798 \\ \hline
Speech    &  0.843/0.841 & 0.827/0.826 & 0.805/0.810 \\ \hline
Sports    &  0.826/0.781 & 0.789/0.778 & 0.749/0.729 \\
\bottomrule
\end{tabular}
\caption{Per category MOS correlations for SDR videos.}
\label{tab:mos_correlations_per_category}
\end{table}

\begin{figure*} 
\centering
\includegraphics[width=1\linewidth, trim={0cm 0cm 0cm 0cm}, clip]{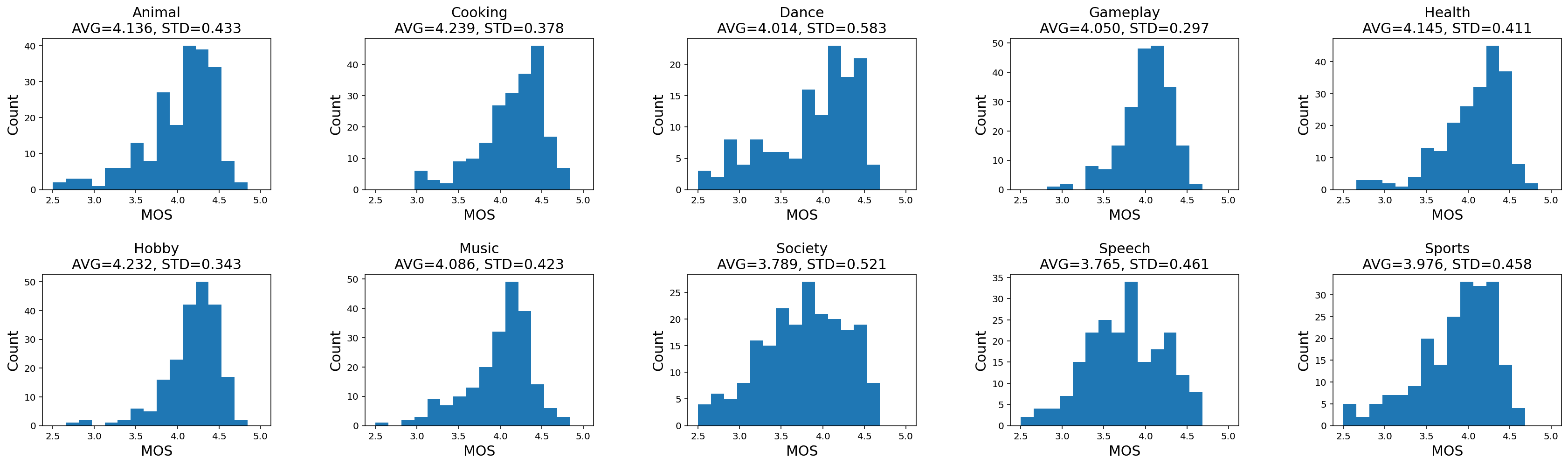}
\caption{MOS distributions per content category for native SDR videos.}
\label{fig:mos_sdr_per_category}
\end{figure*}

\begin{figure} 
\centering
\begin{subfigure}{0.3\linewidth}
\centering
\includegraphics[width=1\linewidth, trim={0cm 0cm 0cm 0cm}, clip]{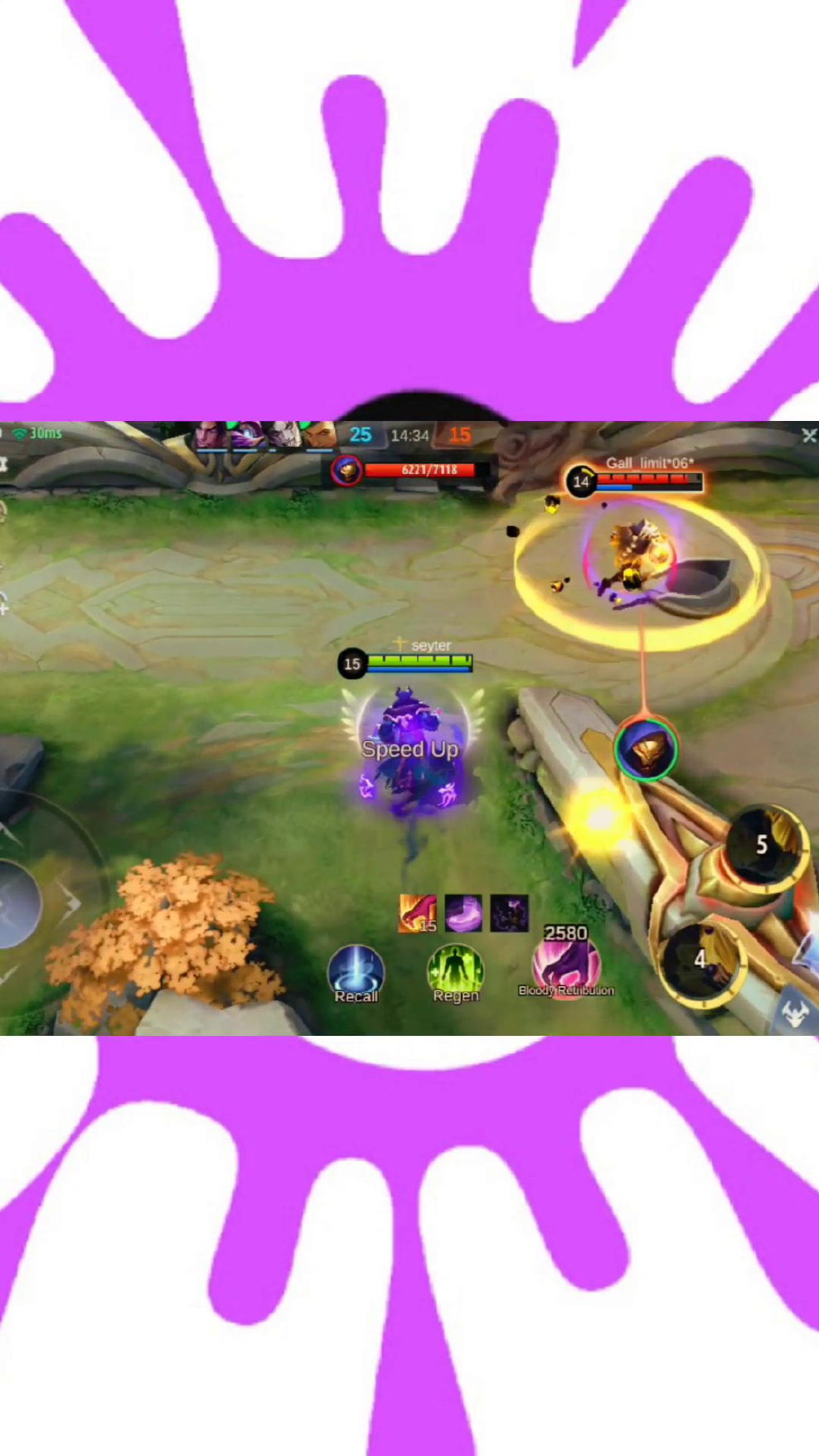}
\captionsetup{font=small, labelformat=empty, justification=centering}
\caption{SDR\_Gameplay\_s0pc MOS=4.26, DOVER=3.26, FAST-VQA=2.95, FasterVQA=3.02}
\end{subfigure}
\begin{subfigure}{0.3\linewidth}
\centering
\includegraphics[width=1\linewidth, trim={0cm 0cm 0cm 0cm}, clip]{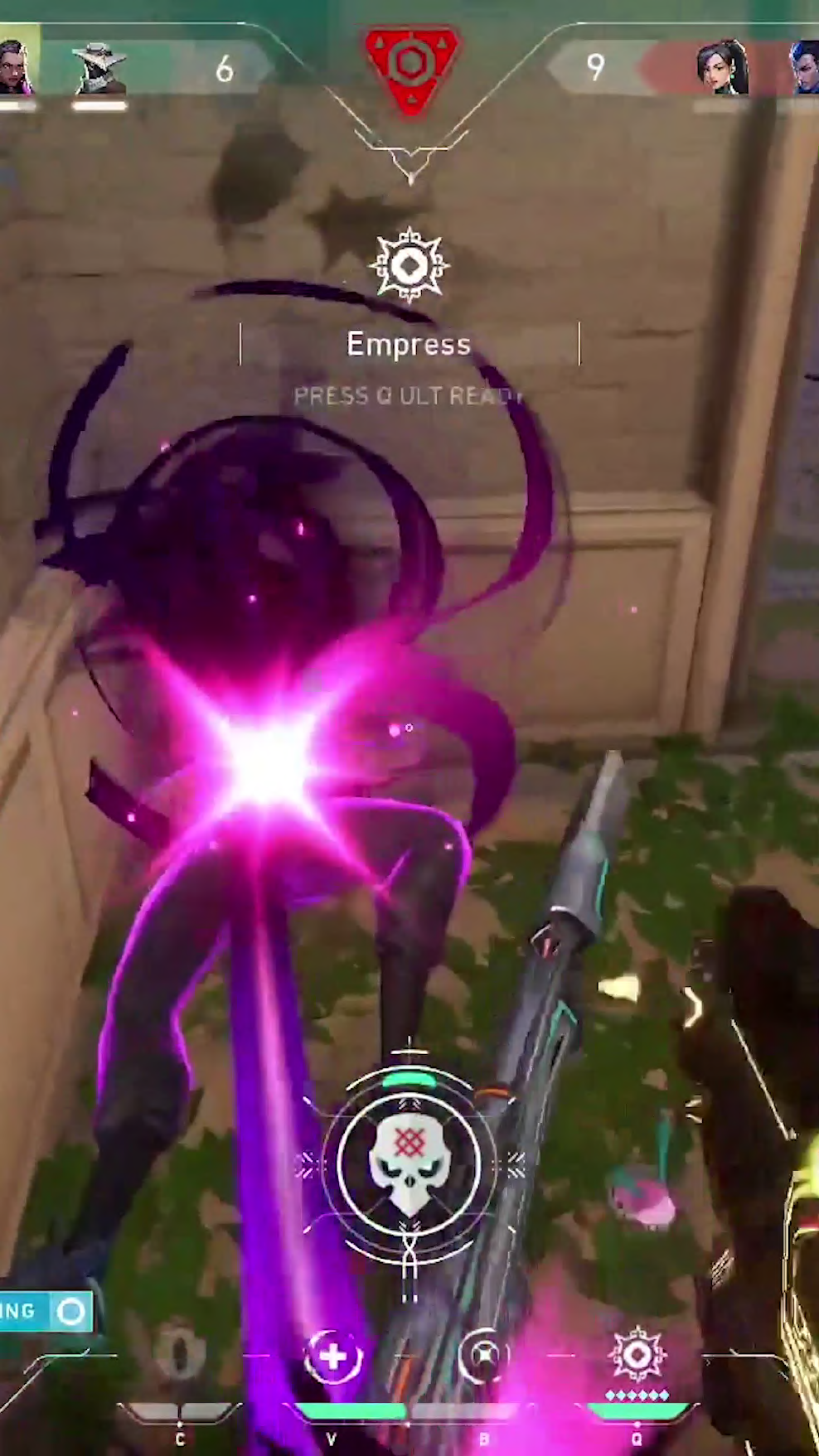}
\captionsetup{font=small, labelformat=empty, justification=centering}
\caption{SDR\_Gameplay\_wcq2 MOS=4.15, DOVER=2.31, FAST-VQA=1.89, FasterVQA=2.36}
\end{subfigure}
\begin{subfigure}{0.3\linewidth}
\centering
\includegraphics[width=1\linewidth, trim={0cm 0cm 0cm 0cm}, clip]{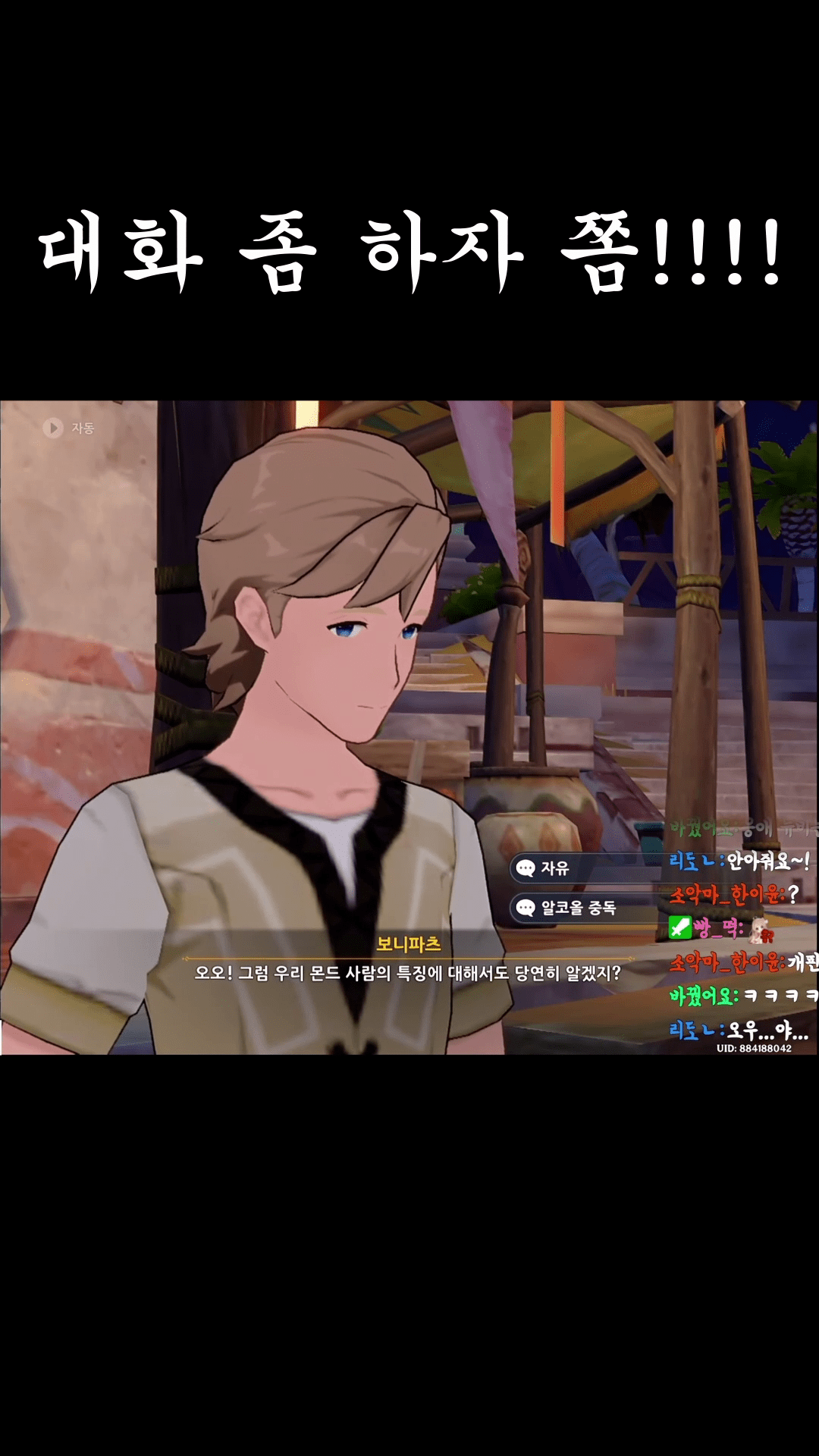}
\captionsetup{font=small, labelformat=empty, justification=centering}
\caption{SDR\_Gameplay\_z5fz MOS=4.22, DOVER=3.20, FAST-VQA=3.23, FasterVQA=2.58}
\end{subfigure}
\caption{Examples of Gameplay SFV with high MOS but low objective metrics.}
\label{fig:high_mos_low_metrics}
\end{figure}

Table~\ref{tab:mos_correlations_per_category} shows the MOS correlations for individual content categories. We found that the highest PLCC for Gameplay is 0.64, significantly lower than other categories. Fig.~\ref{fig:high_mos_low_metrics} showed some examples with high MOS but low metrics scores. We can see that the discrepancy between MOS and metric scores is significant, which implies existing VQA models may need to be retrained on more Gameplay SFV to align with human opinions. 

Fig.~\ref{fig:mos_vs_metrics} shows the scatter plots between MOS and objective metrics. We can see that predicted quality scores are lower than actual MOS in general, and RMSEs are relatively high. It means these models need to be calibrated to get accurate absolute quality scores. 
We also observed some common (not SFV-specific) difficult cases for VQA model, like dark scene and minecraft-style videos. These failure samples can be used to refine existing VQA models.

%% file: conclusion.tex
\section{Conclusion}
\label{sec:Conclusion}
We introduced the YouTube SFV+HDR quality dataset in this paper, which is the first large-scale dataset focusing on SFV and HDR quality with subjective quality labels. A general sampling framework was proposed to maximize the representativeness of videos. We compared the subjective opinions for SFV in three different color conditions (SDR, HDR2SDR, and HDR), and demonstrated the MOS variances among different content categories. We also evaluated the performance of SOTA UGC quality models on SFV, and discussed potential improvements. We hope this work brings new insights and facilitate more research in this area.

\begin{figure}
\centering
\includegraphics[width=0.45\linewidth, trim={0cm 0cm 0cm 0cm}, clip]{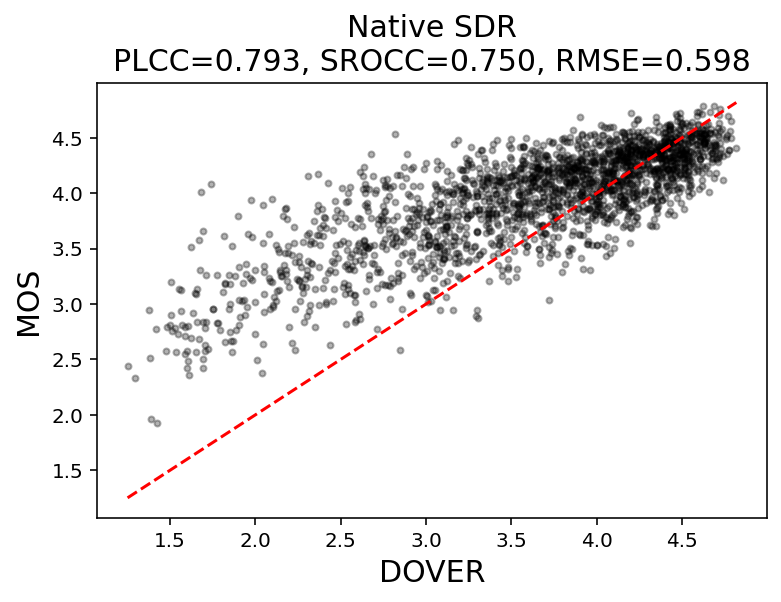}
\includegraphics[width=0.45\linewidth, trim={0cm 0cm 0cm 0cm}, clip]{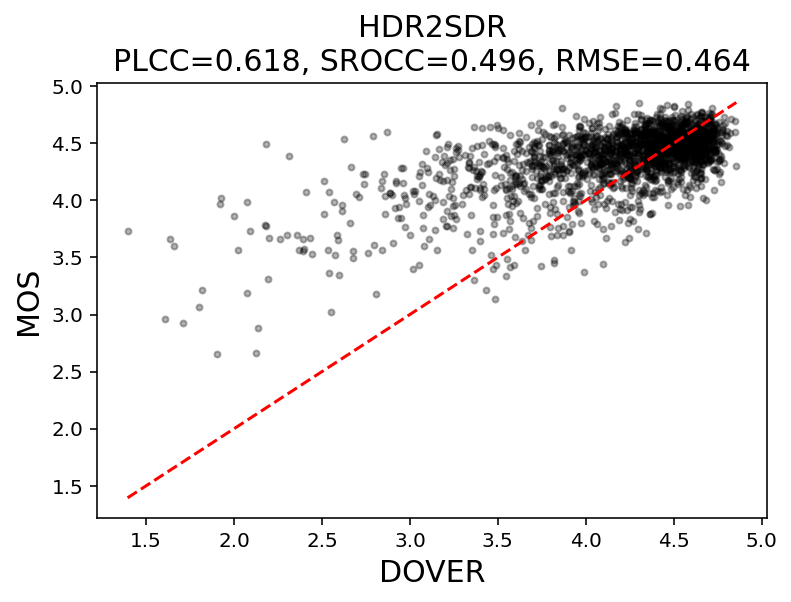}\\
\includegraphics[width=0.45\linewidth, trim={0cm 0cm 0cm 0cm}, clip]{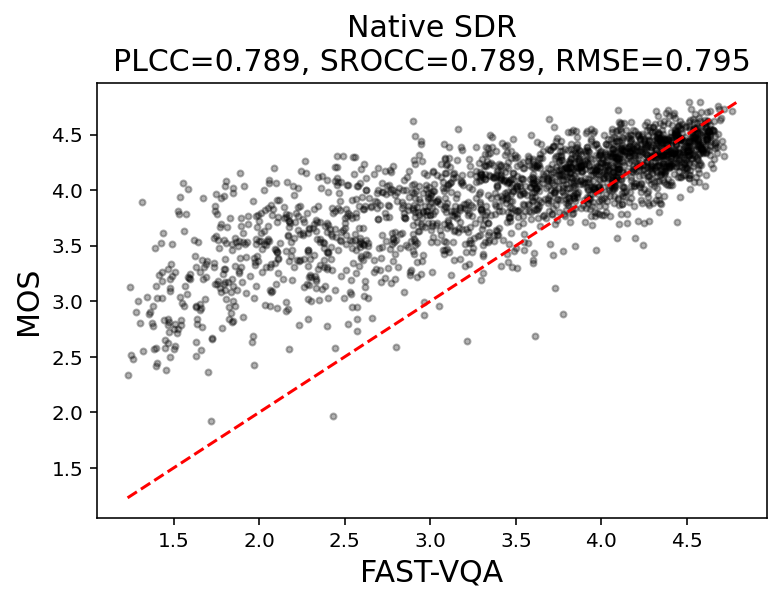}
\includegraphics[width=0.45\linewidth, trim={0cm 0cm 0cm 0cm}, clip]{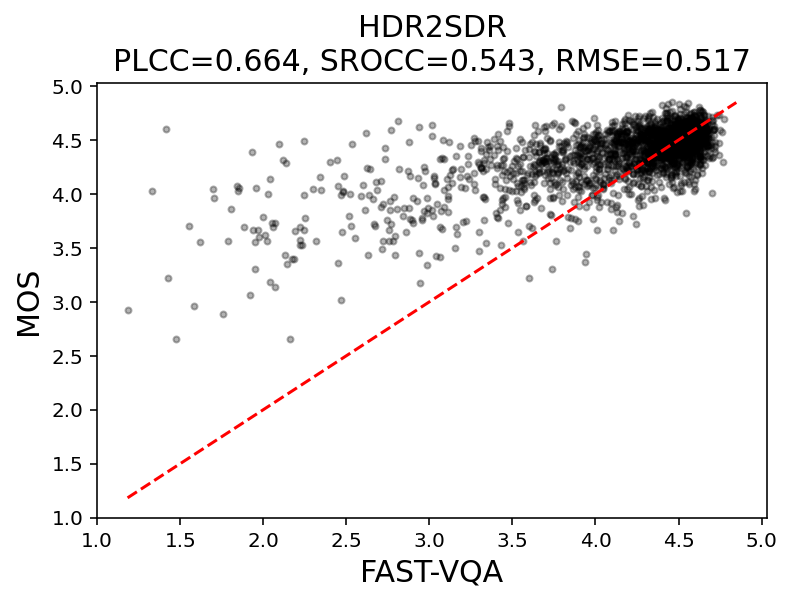}\\
\includegraphics[width=0.45\linewidth, trim={0cm 0cm 0cm 0cm}, clip]{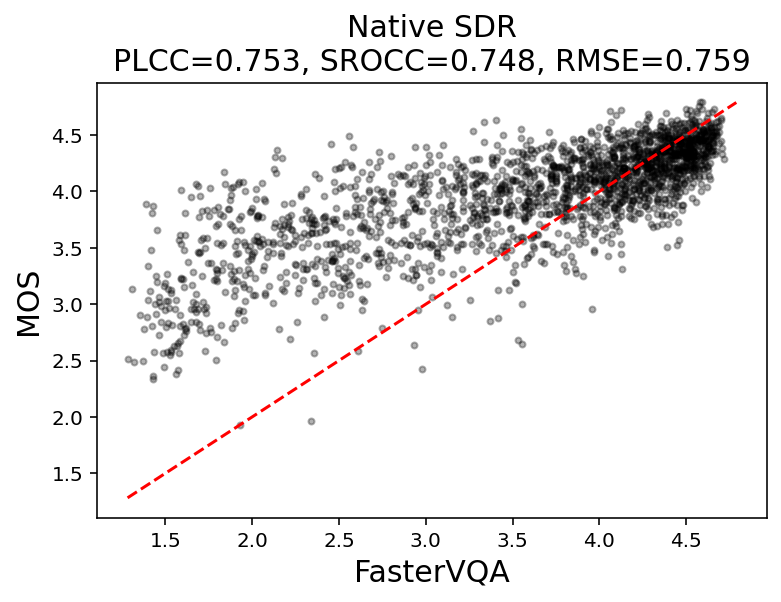}
\includegraphics[width=0.45\linewidth, trim={0cm 0cm 0cm 0cm}, clip]{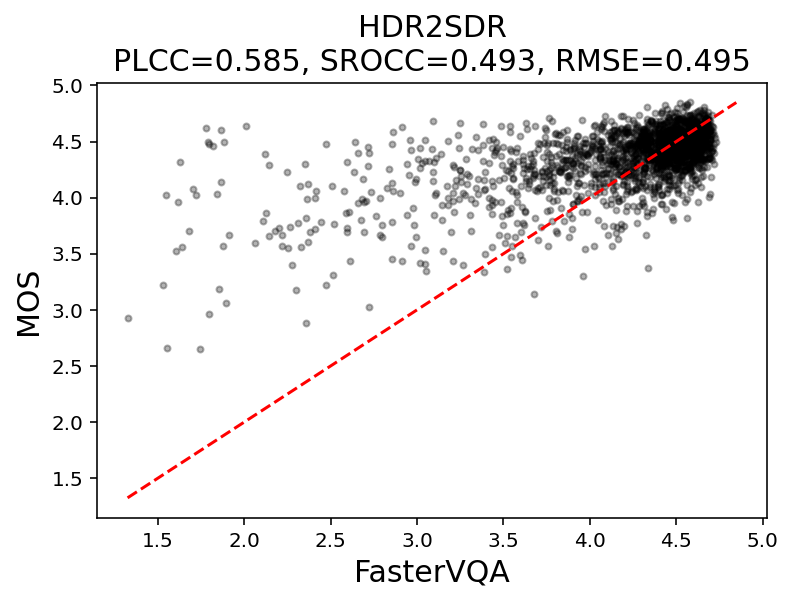}\\
\caption{Scatter plots for native SDR (left column) and HDR converted SDR (right column) v.s. objective metrics DOVER (top), FAST-VQA (middle), and FasterVQA (bottom).}
\label{fig:mos_vs_metrics}
\end{figure}

% \begin{figure} 
% \centering
% \begin{subfigure}{0.3\linewidth}
% \centering
% \includegraphics[width=1\linewidth, trim={0cm 0cm 0cm 0cm}, clip]{figures/SDR_Gameplay_s0pc_0016.png}
% \captionsetup{font=small, labelformat=empty, justification=centering}
% \caption{SDR\_Gameplay\_s0pc MOS=4.26, DOVER=3.26, FAST-VQA=2.95, FasterVQA=3.02}
% \end{subfigure}
% \begin{subfigure}{0.3\linewidth}
% \centering
% \includegraphics[width=1\linewidth, trim={0cm 0cm 0cm 0cm}, clip]{figures/SDR_Gameplay_wcq2_0013.png}
% \captionsetup{font=small, labelformat=empty, justification=centering}
% \caption{SDR\_Gameplay\_wcq2 MOS=4.15, DOVER=2.31, FAST-VQA=1.89, FasterVQA=2.36}
% \end{subfigure}
% \begin{subfigure}{0.3\linewidth}
% \centering
% \includegraphics[width=1\linewidth, trim={0cm 0cm 0cm 0cm}, clip]{figures/SDR_Gameplay_z5fz_0004.png}
% \captionsetup{font=small, labelformat=empty, justification=centering}
% \caption{SDR\_Gameplay\_z5fz MOS=4.22, DOVER=3.20, FAST-VQA=3.23, FasterVQA=2.58}
% \end{subfigure}
% \caption{Examples of Gameplay Shorts with high MOS but low objective metrics.}
% \label{fig:high_mos_low_metrics}
% \end{figure}